%% file: PaperForReview.tex
\documentclass[10pt,twocolumn,letterpaper]{article}

\usepackage[pagenumbers]{cvpr} 
\usepackage[table]{xcolor}
\usepackage{graphicx}
\usepackage{amsmath}
\usepackage{amssymb}
\usepackage{booktabs}
\usepackage{comment}
\usepackage{wrapfig}
\usepackage{enumitem}
\usepackage{multirow}
\usepackage{siunitx}
\usepackage[all]{nowidow}
\usepackage{bm}
\usepackage[ruled,linesnumbered]{algorithm2e}
\usepackage{tabularx}
\usepackage{arydshln}

\newcommand{\settitle}{\maketitle}

\newcommand{\ruofan}{}

\newcommand{\model}{SPIDR}
\newcommand{\citep}{\cite}
\newcommand{\cites}{\cite}
\newcommand{\authorname}{\text}

\usepackage[pagebackref,breaklinks,colorlinks]{hyperref}

\usepackage[capitalize]{cleveref}
\crefname{section}{Sec.}{Secs.}
\Crefname{section}{Section}{Sections}
\Crefname{table}{Table}{Tables}
\crefname{table}{Tab.}{Tabs.}

\def\cvprPaperID{8} 
\def\confName{X}
\def\confYear{2023}

\begin{document}

\title{\emph{\model}: SDF-based Neural Point Fields for Illumination and Deformation}


\newcommand*{\affaddr}[1]{#1} 
\newcommand*{\affmark}[1][*]{\textsuperscript{#1}}
\newcommand*{\email}[1]{\texttt{#1}}

\author{Ruofan Liang\affmark[1,4]\quad Jiahao Zhang\affmark[1]\quad
Haoda Li\affmark[1,2]\quad Chen Yang\affmark[3]\quad Yushi Guan\affmark[1]\quad Nandita Vijaykumar\affmark[1,4]\\[5pt]
\affaddr{
\affmark[1]University of Toronto\;
\affmark[2]UC Berkeley\;
\affmark[3]Shanghai Jiao Tong University\;
\affmark[4]Vector Institute
}\\[1pt]
}


\twocolumn[{%
\renewcommand\twocolumn[1][]{#1}%
\settitle
\begin{center}
    \centering
    \small
    \vspace{-15pt}
    \includegraphics[width=1.\textwidth,trim={0cm 0cm 0cm 0.5cm},clip]{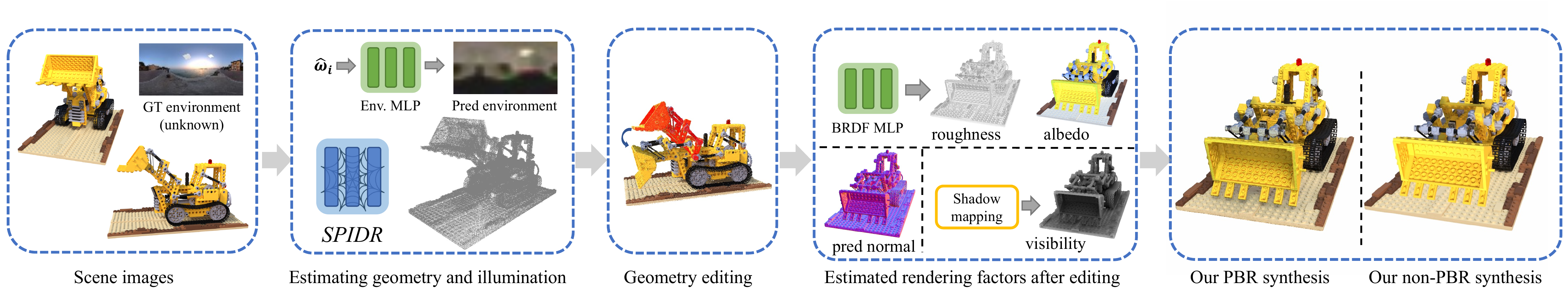}
    \vspace*{-15pt}
    \captionof{figure}{\small
    \textbf{Overview of SPIDR}.
    Given a set of scene images under an unknown illumination, \model\ uses a hybrid neural implicit point representation to learn the scene geometry, radiance, and BRDF parameters. \model\ also employs an MLP model to learn and represent environment illumination. 
    After obtaining a trained \model\ model, users can perform various geometry editing using our explicit point cloud representation. 
    \model\ then updates estimated rendering factors based on the user's geometry editing.
    \model\ finally uses estimated rendering factors to synthesize the deformed object image using physically-based rendering (PBR) with updated lighting and shadows. 
    }
    \label{fig:demo}
\end{center}%
}]
\begin{abstract}
\input{abs}
Project webpage: \url{https://nexuslrf.github.io/SPIDR_webpage}.
\end{abstract}

\input{introduction}
\input{related_work}
\input{preliminaries}
\input{method}
\input{method_light}

\input{method_deform}
\input{experiments}

\input{conclusions}

{\small
\bibliographystyle{ieee_fullname}
\bibliography{ref}
}

\clearpage
\section*{Supplementary}
\input{appendix.tex}
\end{document}

%% file: abs.tex


Neural radiance fields (NeRFs) have recently emerged as a promising approach for 3D reconstruction and novel view synthesis. 
However, NeRF-based methods encode shape, reflectance, and illumination \emph{implicitly} and this makes it challenging for users to manipulate these properties in the rendered images explicitly.
Existing approaches only enable limited editing of the scene and deformation of the geometry. Furthermore, no existing work enables accurate scene illumination after object deformation. 
In this work, we introduce \model, a new hybrid neural SDF representation. \model\  combines point cloud and neural implicit representations to enable the reconstruction of higher quality object surfaces for geometry deformation and lighting estimation. 
To more accurately capture environment illumination for scene relighting, we propose a novel neural implicit model to learn  environment light. To enable more accurate illumination updates after deformation, we use the shadow mapping technique to approximate the light visibility updates caused by geometry editing.
We demonstrate the effectiveness of \model\ in enabling high quality geometry editing with more accurate updates to the illumination of the scene.

%% file: introduction.tex
\vspace{-10pt}
\section{Introduction}
The recent advancements in neural radiance fields (NeRF) \citep{mildenhall2020nerf}
have demonstrated exciting new capabilities in scene reconstruction and novel view synthesis. 
Recent research has investigated various aspects of implicit representations including training/inference acceleration, generative synthesis, modeling dynamic scenes, and relighting \citep{dellaert2020neural}.
An important and desirable aspect of 3D representations is their ability to be easily edited for applications such as {3D content production, games, and the movie industry}.
However, the geometry and illumination of objects/scenes represented by NeRFs are fundamentally challenging to edit for two reasons.
First, NeRF-based methods implicitly represent objects and scenes with neural functions. Thus explicit geometry manipulations such as fine-grained rigid/non-rigid body deformations on the NeRF representations are challenging.
Second, the realistic editing of scenes often requires a corresponding change in the illumination of the object/scene. For example, unshadowed regions become shadowed if light occlusion happens after the deformation. Since illumination parameters (e.g., environment and materials) are not explicitly modeled or represented in most NeRF-based methods, achieving accurate illumination after deformation is challenging. 

Recent work that aim to improve the editability of NeRFs \citep{Yuan22NeRFEditing, neumesh, xu2022deforming} involve
extracting mesh representations. However, these approaches do not tackle the scene illumination challenge posed by geometry deformation. Recent research has also investigated various aspects of illuminations \citep{nerv2021, nerfactor, munkberg2021extracting}, including scene relighting, material editing, and environmental light estimation. However, these approaches cannot be directly applied to illumination
changes due to object deformations. This is because object deformation may cause occlusions in the scene, and to derive these effects, we need to update the estimated surface light visibility map. This is a challenging task and current relighting methods do not enable updates to the visibility map. Furthermore, existing approaches do not provide sufficiently accurate environment light estimation for synthesizing clear shadow changes. Thus, enabling both geometry deformation and illumination is a challenging and yet unsolved task.

Our \textbf{goal} in this work is twofold. First, we aim to enable the {fine-grained} geometry deformation in NeRFs, including {rigid body movement and non-rigid body deformation}. Second, we aim to enable illumination/shadow changes during the object deformations. 
To achieve these goals, we present \model, an SDF-based hybrid NeRF model.
\model\  is designed based on two major ideas:
\textbf{(1) SDF-based hybrid point cloud representation}: \emph{enabling better editability.} {We propose a hybrid model that is a combination of point cloud and neural implicit representations and is designed to estimate the signed distance field (SDF).} This hybrid representation enables the manipulation of the object geometry in two ways: 
mesh-guided deformations and direct point cloud manipulations for non-rigid and rigid body transformations, respectively.
{Unlike SDF-based NeRF MLPs \citep{yariv2021volume, wang2021neus}, the point cloud representation enables  direct manipulation of the object geometry to achieve the rigid-body transformation.
}
\textbf{(2) Explicit estimation of environment light and visibility}: \emph{enabling better illumination with deformation.} To estimate environment light more accurately, we employ an implicit coordinate-based MLP which {takes in an incident light direction and predicts the corresponding light intensity}. %
To enable illumination updates after deformation, unlike prior work that use neural models \citep{nerv2021, nerfactor}, 
we employ the shadow mapping technique to efficiently approximate light visibility with NeRF's estimated depth maps.
\\\noindent
In summary, our contributions are as follows. We propose:
{
\begin{itemize}[leftmargin=*, topsep=0pt, itemsep=0pt, partopsep=0pt, parsep=1pt]
\item The first NeRF-based editing approach that enables both  geometry deformation and illumination/shadow changes from deformation.
\item An SDF-based hybrid implicit representation with  new regularizations that enables the reconstruction of high quality geometry for surface illumination. 
\item The use of a coordinate-based MLP to represent the environment light, which enables {more accurate} estimation of lighting that is required for illumination changes.
\item The employment of the shadow mapping technique to approximate visibility for the rendering of shadowing effects after object deformation. 
\end{itemize}
}
\noindent
{Our experiments demonstrate that \model\ can obtain higher quality reconstructed object surfaces and more accurate estimations of the environment light. These essential improvements enable \model\ to render deformed scenes with updated illumination and shadowing effects. 
}

%% file: related_work.tex
\begin{table*}[h]
    \centering
    \small
    \setlength\tabcolsep{1pt}
        \begin{tabularx}{\linewidth}%
        {>{\centering\arraybackslash}p{0.12\linewidth}*{5}{>{\centering\arraybackslash}X}p{1em}}
\multicolumn{7}{c}{\includegraphics[width=1\linewidth,trim={0pt 40pt 0 2pt},clip]{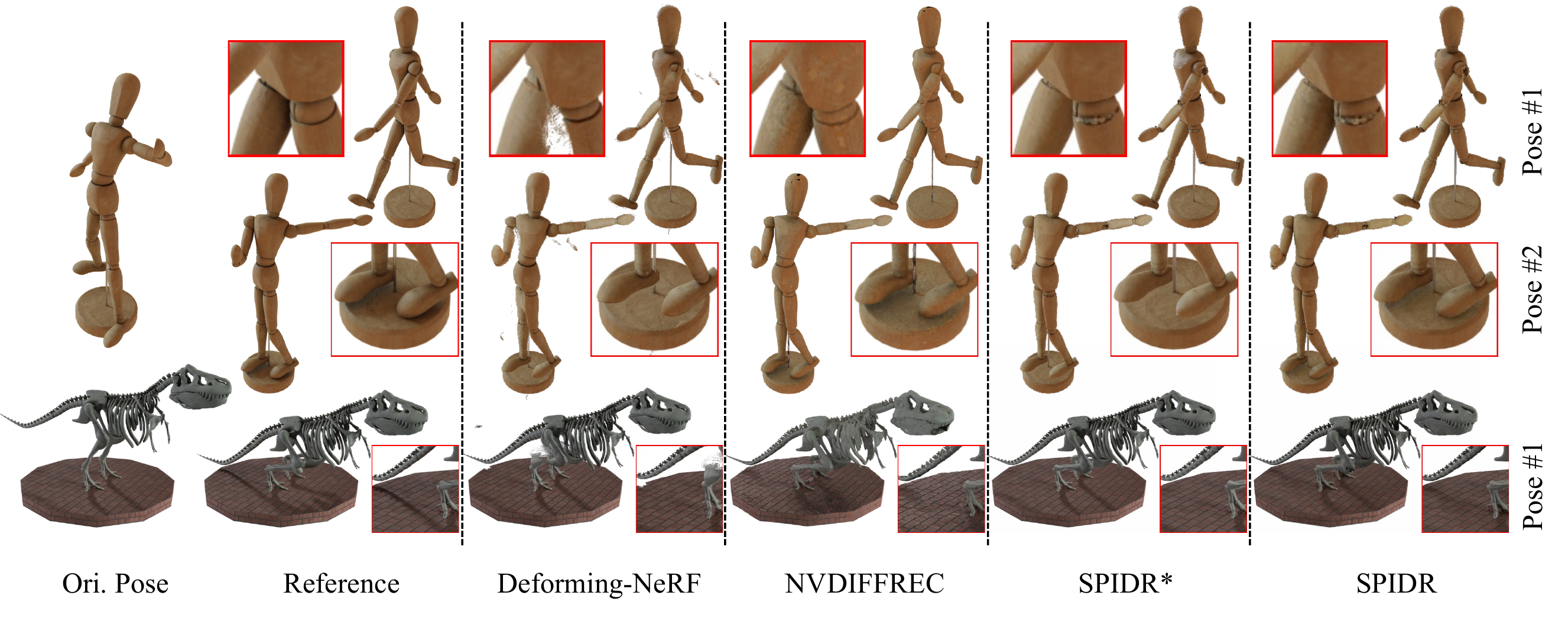}}
\\
Ori. Pose & Reference & Deforming-NeRF \cite{xu2022deforming} & NVDiffRec \cite{munkberg2021extracting} & \model$^*$ & \model &
        \end{tabularx}%
    \vspace{-5pt}
    \makeatletter\def\@captype{figure}\makeatother
    \caption{Qualitative comparison on rendering deformed scenes. We use the rigged cages to guide the deforming-NeRF's deformations. NVDIFFREC's meshes are directly rigged with ground truth bones. \model$^*$ and \model\ use the deformed ground truth mesh to guide the deformation. NVDIFFREC uses Blender Cycles for rendering, while the other methods use neural models for rendering.
    \label{fig:deform_compare}}
\vspace{-10pt}
\end{table*}


\section{Related Work}
\textbf{Neural rendering and scene representation.} Neural rendering is a class of reconstruction and rendering approaches that use deep networks to learn complex mappings from captured images to novel images \citep{tewari2020state}.
Neural radiance field (NeRF) \citep{mildenhall2020nerf} is one representative work that shows how the current state-of-the-art neural rendering methods \citep{barron2021mip, barron2022mip, verbin2021ref} combine volume rendering and neural network based implicit representation for photo-realistic novel view synthesis. Follow-up works further improve the quality of reconstructed geometry \citep{yariv2021volume, wang2021neus} using SDFs. 
For more efficient training and rendering, hybrid neural representations that combine implicit MLP and other discrete spatial representations such as voxel grids \citep{liu2020neural, fridovich2022plenoxels}, point clouds \citep{ost2022neural, xu2022point}, and hash tables \citep{mueller2022instant} have been proposed. These approaches however do not enable high quality scene editing and scene illumination after deformation. 

\textbf{Geometry editing and deformation.}
A large body of prior research investigates geometry modeling and editing for explicit 3D representations (e.g., meshes) \citep{kholgade20143d, jacobson2012fast, yifan2020neural}. These approaches cannot be directly applied to neural implicit representations. 
For neural representations, scene composition has been investigated by several works \citep{yang2021learning, tang2022compressible, lazova2022control}, where objects can be added to or moved in the scene, but the shape of objects cannot be changed.
Recent works also explore ways to achieve user-defined geometry deformation. \authorname{Liu \etal} \citep{liu2021editing} allows shape editing with the user's scribble but is limited to simple objects belonging to certain categories. 
{Closest to our approach are \citep{Yuan22NeRFEditing, neumesh, xu2022deforming}, where they achieve object deformations by using mesh-guided deformation. {In addition to mesh-guided deformation, 
our method also supports direct point could manipulation, which can achieve more realistic rigid-body deformations.}} 
None of these works address the illumination changes caused by deformation. 
\authorname{Guo \etal} \citep{guo2020object} enables relighting when composing scenes with multiple objects. This approach however does not enable shape editing nor the resulting change in illumination. 

\textbf{Lighting estimation.}
Lighting estimation, also known as inverse rendering \citep{marschner1998inverse}, is a long-existing problem that aims to estimate surface reflectance properties and lighting conditions from  images. Prior works have shown how to obtain accurate BRDF properties under known lighting conditions \citep{matusik2003data,aittala2016reflectance,deschaintre2018single}; and how to estimate environment light from objects with known geometry \citep{richter2016instant, legendre2019deeplight, park2020seeing}.
Recent works also leverage NeRF method to learn scene lighting and material reflectance. 
\authorname{Bi \etal} \citep{bi2020neural} attempt to learn the material reflectance properties with NeRF, but both methods require known lighting conditions.
Some recent work jointly estimate environment light and reflectance with images under unknown lighting conditions \citep{boss2021nerd, boss2021neural, zhang2021physg, munkberg2021extracting}, 
but their lighting approximation methods (e.g., spherical Gaussian, pre-integrated maps) do not consider self-occlusions, thus cannot directly render shadowing effects caused by occlusion.
NeRV \citep{nerv2021} and NeRFactor \citep{nerfactor} have explicit light visibility estimation in their model, allowing them to render relightable shadowing effects. However, the rendered results of these methods are over-smoothed and lack high-frequency details.
Additionally, these NeRF-based lighting estimation methods mainly focus on relighting static scenes and cannot be applied to address the illumination changes caused by geometry editing. 
In contrast, our model is able to render updated lighting and shadowing effects after editing the geometry of a NeRF scene.

%% file: preliminaries.tex
\vspace{-6pt}
\section{Preliminaries}

\textbf{MVS and Point-based NeRF.}\label{sec:pointnerf}
 Several works leverage multi-view stereo (MVS) methods to provide geometry and appearance priors for NeRF, enabling efficient training and inference \citep{chen2021mvsnerf, lin2021efficient, xu2022point}. 
Point-NeRF\citep{xu2022point} is a hybrid NeRF representation with the point cloud representing explicit geometry.
The point cloud used by Point-NeRF is first initialized by MVS methods \citep{schonberger2016pixelwise, yao2018mvsnet}.
This neural point cloud is formulated as $\mathcal P = \{(\mathbf p_i, \mathbf f_i)\}$, where $\mathbf p_i$ and $\mathbf{f}_i$ denote point position and point feature (from CNN extractors), respectively. 
During volume rendering, 
the volume density $\sigma({\mathbf{x}})$ of a query point $\mathbf x$ along the ray with direction
$\hat{\mathbf{v}}$ is an inverse distance interpolation of density values of the neighboring neural points.
The density of each neural point conditioned on query point $\mathbf x$ is predicted using a PointNet-like MLP \citep{qi2017pointnet} model $F$ followed by a spatial MLP $T$: 
\begin{align}
\vspace{-10pt}
    \sigma({\mathbf{x}}) = \sum_i \frac{w_{\mathbf{p}_i}}{\sum w_{\mathbf{p}_i}}T(\mathbf{f}_{i, \mathbf{x}}),\quad\mathbf{f}_{i, \mathbf{x}} = F(\mathbf{f}_i,\ \mathbf{x} - \mathbf{p_i})
\end{align}
where $w_{\mathbf{p}_i} = {\|\mathbf{x} - \mathbf{p}_i\|}^{-1}$, denoting the inverse distance. The view-dependent radiance $\mathbf c$ is predicted by a radiance MLP $R$ with interpolated point features and view direction $\hat{\mathbf{v}}$ as the input:
\vspace{-8pt}
\begin{equation}
    \vspace{-6pt}
    \mathbf{c}({\mathbf x}) = R(\mathbf{f}_{\mathbf{x}}, \hat{\mathbf{v}}),\quad\mathbf{f}_{\mathbf{x}} = \sum_i \frac{w_{\mathbf{p}_i}}{\sum w_{\mathbf{p}_i}} \mathbf{f}_{i, \mathbf{x}} \label{eq:interp_feat}
\end{equation}
Then the radiance of sampled points $\mathbf x_t$ along the ray $\mathbf r$ is accumulated following NeRF's volume rendering equation:
\vspace{-10pt}
\begin{align}
\vspace{-10pt}
    {\mathbf C}(\mathbf r) &= \sum_{t} w_t \mathbf c(\mathbf{x}_t) \notag \\
    \text{with } w_t = \exp(-\sum_{j=1}^{t-1}\sigma&(\mathbf{x}_t) \delta_t)(1 - \exp(-\sigma(\mathbf{x}_t) \delta_t) ) \label{eq_radiance}
\end{align}
where $\delta_t$ denotes the distance between adjacent sampled positions along the ray.

\textbf{Implicit neural surface.} 
Recent works show that implicit surface representation can improve the quality of NeRF's reconstructed geometry \citep{oechsle2021unisurf, yariv2021volume,wang2021neus}. \authorname{Yariv \etal} \citep{yariv2021volume} illustrate a concrete mathematical conversion of volume density $\sigma(\mathbf{x})$ from predicted SDF $d_{\Omega}(\mathbf{x})$ of the space $\varOmega \subset \mathbb{R}^3$ where the target object locates: 
\vspace{-10pt}
\begin{equation}
\vspace{-6pt}
    \sigma(\mathbf{x})
    =  \frac{1}{\beta} \Psi_\beta (-d_{\varOmega}(\mathbf{x}))
    \label{sdf_trans}
\end{equation}
$\beta$ is a learnable parameter, and $\Psi_\beta$ is the cumulative distribution function (CDF) of the Laplace distribution:
\vspace{-5pt}
\begin{equation}
    \Psi_\beta(s) = \begin{cases}
\frac{1}{2}\exp(\frac{s}{\beta}) & \text{if } s \leq 0,\\
1 - \frac{1}{2}\exp(-\frac{s}{\beta}) & \text{if } s > 0.
\end{cases}
\vspace{-6pt}
\end{equation}
Following this parameterization, we get a high-quality smooth surface from the zero level-set of predicted SDF values, and surface normal can be computed as the gradient of the predicted SDF w.r.t. surface point $\mathbf x$, i.e., $
    \hat{\mathbf{n}}_{\mathbf{x}} = \frac{\nabla d_\varOmega(\mathbf x)}{\|\nabla d_\varOmega(\mathbf x)\|}$. 

\textbf{The rendering equation.} 
Mathematically, the outgoing radiance of a surface point $\mathbf{x}$ with normal $\hat{\mathbf{n}}$ from outgoing direction $\hat{\bm{\omega}}_o$ can be described by the physically-based rendering equation (PBR) \citep{kajiya1986rendering}:
\vspace{-5pt}
\begin{equation}
\vspace{-5pt}
    L_o(\mathbf{x}, \hat{\bm{\omega}}_o) = \int_{\Omega}  L_i(\mathbf{x},  \hat{\bm{\omega}}_i) f_r(\mathbf{x}, \hat{\bm{\omega}}_i, \hat{\bm{\omega}}_o) (\hat{\mathbf{n}}\cdot \hat{\bm{\omega}}_i)  d \hat{\bm{\omega}}_i \label{eq:re}
\end{equation}
where $\hat{\bm{\omega}}_i$ denotes incoming light direction, $\Omega$ denotes the hemisphere centered at $\hat{\mathbf{n}}$, $L_i(\mathbf{x},  \hat{\bm{\omega}}_i)$ is the incoming radiance of $\mathbf{x}$ from $\hat{\bm{\omega}}_i$, $f_r$ is the bidirectional reflectance distribution function (BRDF) that describes the portion of reflected radiance at direction $\hat{\bm{\omega}}_o$ from direction $\hat{\bm{\omega}}_i$. 

%% file: method.tex

\section{SDF-based Point-NeRF}\label{sec:sdf_pnerf}
This section describes how our hybrid NeRF model is parameterized for learning scene geometry and radiance.
Since our objectives rely on high-quality geometry for operations such as mesh extraction and shadow mapping, the original NeRF's reconstructed geometry is insufficient for these operations. 
Since we use a discrete hybrid model instead of a continuous MLP model, extra regularizations on SDF predictions are required. 


\subsection{Parameterization}

The structure of our model generally follows the design of Point-NeRF \citep{xu2022point} described in Sec. \ref{sec:pointnerf}.
To incorporate SDF-based NeRF rendering, we make the density MLP $T$ to predict SDF value $d$ instead of volume density $\sigma$. These implicit SDF values $d$ can then be converted into density values using equation \ref{sdf_trans}. The geometry estimation MLP in our model now becomes  
\begin{align}
    d({\mathbf{x}}) = \sum_i  \frac{w_{\mathbf{p}_i}}{\sum w_{\mathbf{p}_i}}d_i,\quad d_i = T(\mathbf{f}_{i, \mathbf{x}}) 
\end{align}
To predict radiance, our model outputs both view-independent diffuse color and view-dependent specular color. This decomposition makes it easier for the model to learn view-independent color (Section \ref{sec:lighting}).
The diffuse color $\mathbf{c}_{d}$ is obtained by feeding the interpolated point feature $\mathbf{f}_{\mathbf{x}}$ (Eqn. \ref{eq:interp_feat}) to the diffuse color MLP $R_d(\mathbf{f}_{\mathbf{x}})$.
Similarly, the specular color $\mathbf{c}_{s}$ is obtained using another MLP branch $R_s(\mathbf{f}_{\mathbf{x}}, \hat{\mathbf{v}}, \hat{\mathbf{n}}_{\mathbf{x}}, 	\hat{\mathbf{v}}\cdot \hat{\mathbf{n}}_{\mathbf{x}})$.
We add $\hat{\mathbf{n}}_{\mathbf{x}}$ and $\hat{\mathbf{v}} \cdot \hat{\mathbf{n}}_{\mathbf{x}}$ as extra inputs to this view-dependent MLP $R_s$. The inner product $\hat{\mathbf{v}} \cdot \hat{\mathbf{n}}_{\mathbf{x}}$ represents cosine value of the angle between normal and view or reflectected direction. $\hat{\mathbf{v}}, \hat{\mathbf{n}}_{\mathbf{x}}$ and $\hat{\mathbf{v}}\cdot \hat{\mathbf{n}}_{\mathbf{x}}$ are necessary terms for calculating specular radiance with BRDF. We expect the output of MLP $R_s$ to be an approximation of the integral in Eqn. \ref{eq:re} with the specular BRDF.

The final radiance $\mathbf{c}(\mathbf x)$ of each sampled point $\mathbf{x}_t$ along the ray that is used in Eqn. \ref{eq_radiance} then becomes
\vspace{-6pt}
\begin{equation}
    \mathbf{c}(\mathbf x) = R_d(\mathbf{f}_{\mathbf{x}}) + R_s(\mathbf{f}_{\mathbf{x}}, \hat{\mathbf{v}}, \hat{\mathbf{n}}_{\mathbf{x}}, 	\hat{\mathbf{v}}\cdot \hat{\mathbf{n}}_{\mathbf{x}})
    \label{eq:d+s}
\end{equation}
\subsection{Normal Representation}\label{sec:point_normal}
Surface normals are frequently used in physically-based rendering, however, computing normals from an MLP's gradient is a costly operation. Thus, we explicitly attach a normal vector $\hat{\mathbf{n}}^\prime_{\mathbf{p}_i}$ to each neural point $\mathbf{p}_i$ in the point cloud. These normal vectors are supervised by the normals from the MLP 's gradients during training. Similar to other per-point attributes, the normal $\hat{\mathbf{n}}^{\prime}_{\mathbf{x}}$ of the sampled point $\mathbf{x}$ is an interpolation of normal vectors of the neighboring points. A simple weighted L2 loss is applied between the interpolated normal $\hat{\mathbf{n}}^{\prime}_{\mathbf{x}}$ and the computed normal $\hat{\mathbf{n}}_{\mathbf{x}}$ for query point $\mathbf{x}$:
\begin{align}
    \mathcal{L}_n = \sum_{\mathbf{x}} w_{\mathbf{x}}\|\hat{\mathbf{n}}_{\mathbf{x}}-\hat{\mathbf{n}}^{\prime}_{\mathbf{x}}\|^2, \quad
    \hat{\mathbf{n}}^{\prime}_{\mathbf{x}} = \sum_i  \frac{w_{\mathbf{p}_i}}{\sum w_{\mathbf{p}_i}}\hat{\mathbf{n}}^\prime_{\mathbf{p}_i} \label{eq:interp_normal}
\end{align}
where $w_{\mathbf{x}}$ is the alpha-compositing weights of sampled point $\mathbf{x}$ along its ray $\mathbf{r}$, as shown in Eqn. \ref{eq_radiance}. 
In addition to facilitating deformation, these explicitly predicted normals can effectively eliminate noise in the directly computed normals \citep{verbin2021ref}.  
The normal vectors used in  the view-dependent MLP $R_s$ will also be replaced with these predicted normals in the inference and later training steps.
\subsection{SDF Regularizations}
Unlike prior work \citep{oechsle2021unisurf, yariv2021volume, wang2021neus} that use a large continuous MLP to implicitly represent SDF values for the whole scene, our model has local neural features attached to the discrete point cloud representation. The discretization of implicit neural representation can cause the loss of continuity of the represented signals \citep{reiser2021kilonerf, fridovich2022plenoxels},
because the predicted signals now are mainly controlled by the local neural features instead of a global MLP.
To address this issue, we add two regularization terms to the predicted SDF values based on SDF's geometric properties.

The first regularization is to stabilize the SDF prediction for points inside the surface. 
We want our model to make stable negative SDF value predictions for the regions inside the surface.
To achieve this goal, we propose a negative sparsity loss for all the sampled query points similar to the Cauchy loss used in \citep{hedman2021baking}.
\vspace{-5pt}
{\small\begin{equation}
    \mathcal{L}_s = \sum_{\mathbf{x}} \log\left(1+ \frac{(1-\beta \sigma(\mathbf{x}))^2}{c^2}\right)
\label{eq:l_s}
\vspace{-8pt}
\end{equation}}\\
where $c$ is a hyperparameter that controls the loss scale, and $\beta$ is the parameter used in Eqn. \ref{sdf_trans}. 
By constraining the SDF value of the internal points, this regularization term can also prevent the model from incorrectly learning internal emitters \citep{verbin2021ref} instead of a solid surface when there
are specular highlights.

\begin{figure}
    \vspace{-8pt}
    \centering
    \includegraphics[width=\linewidth,trim={0 5pt 12pt 0},clip]
    {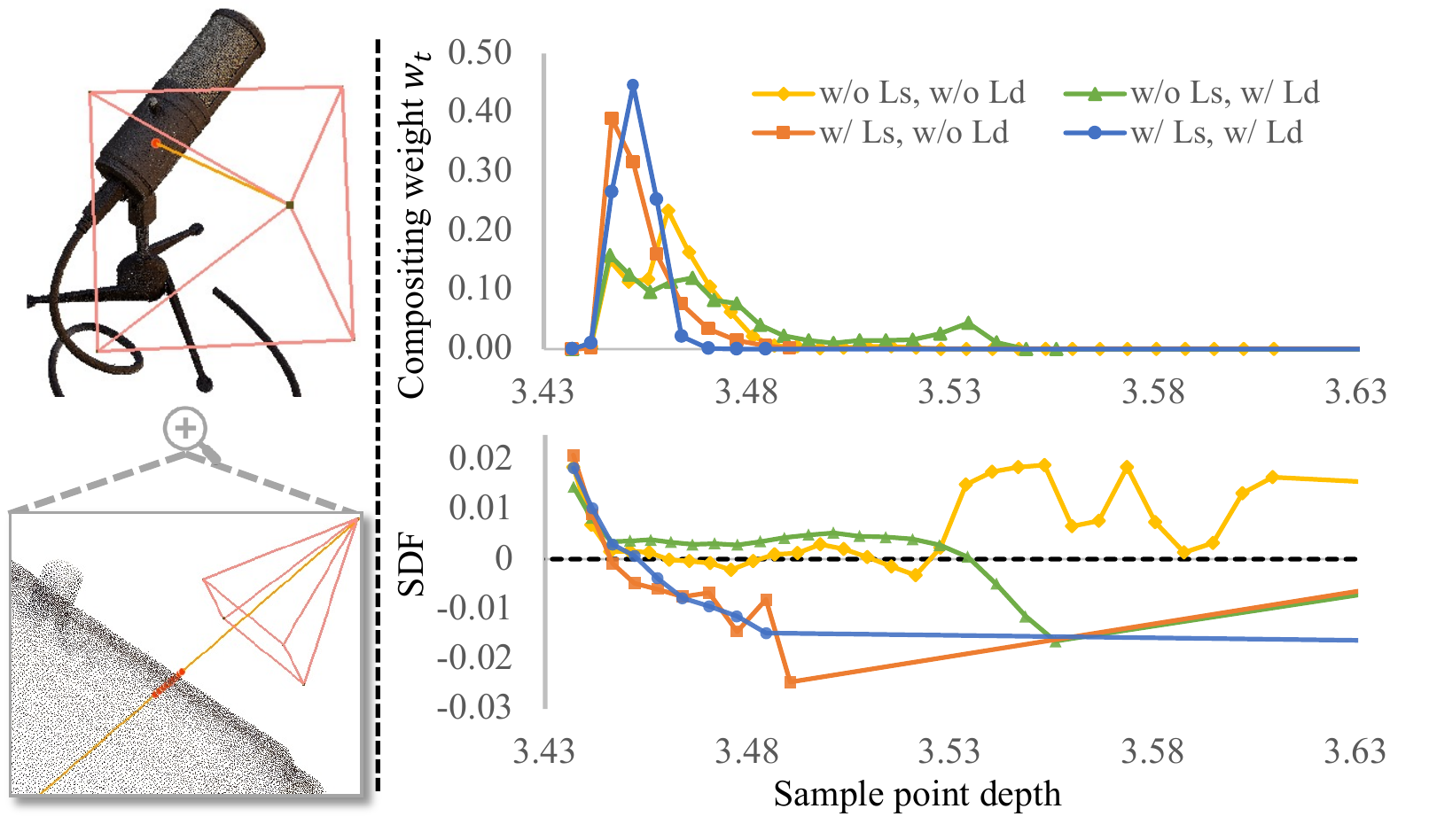}
    \vspace{-15pt}
    \caption{Ablation of our proposed SDF regularizations ($\mathcal{L}_s$ \& $\mathcal{L}_d$) on Mic scene. We plot the curves of SDF and compositing weight $w_t$ along the sampled points shown on the left.}
    \label{fig:ablation_sdf}
    \vspace{-16pt}
\end{figure}

The second regularization is to improve the continuity and consistency of predicted SDF values. 
Since the signed distance field defines the local spatial distance information of 3D positions, we can utilize this property to regularize the predicted SDF values along the sampled rays.
Given two adjacent sampled points $\mathbf{x}_t, \mathbf{x}_{t+1}$ with sufficiently small distance along the same ray $\mathbf{r}_i$, the difference of predicted SDF values $\Delta d_t = d_{\mathbf{x}_{t+1}} - d_{\mathbf{x}_{t}}$ can be approximated by the projection of the relative distance  $\delta_t = \|\mathbf{x}_{t+1} - \mathbf{x}_{t}\|$ 
in the normal direction $\hat{\mathbf{n}}_{\mathbf{x}_{t}}$, which is $\Delta d^\prime_t \approx (\hat{\mathbf{v}}\cdot \hat{\mathbf{n}}_{\mathbf{x}_{t}}) \delta_t$.
We can therefore use the real distance value $\delta_t$ to constrain the changes of predicted SDF $d_{\mathbf{x}}$ with another L2 loss term
\vspace{-5pt}
\begin{equation}
    \mathcal{L}_d = \sum_{t}\|\Delta d^\prime_t - \Delta d_t\|^2
\label{eq:l_d}
\vspace{-5pt}
\end{equation}
This regularization term can help our discrete NeRF model better learn the geometry of the target object without being affected by the discrete representation. 
We show the effectiveness of proposed regularizations in Figure \ref{fig:ablation_sdf}, which will be further discussed in Sec. \ref{sec:ablation}.

\vspace{-4pt}
\subsection{Training}\label{sec:train_s1}
The training process is essentially the same as other NeRF methods, but with the above regularization terms. The full loss function is defined as 
\vspace{-4pt}
\begin{equation}
    \mathcal{L} = \mathcal{L}_{color} + \lambda_n \mathcal{L}_n + \lambda_s \mathcal{L}_s + \lambda_d \mathcal{L}_d 
\vspace{-4pt}
\end{equation}
where $\mathcal{L}_{color}$ is an L1 loss between the ground truth and the rendered images, $\lambda_n, \lambda_s, \lambda_d$ are loss hyperparameters. 
In order to feed stable normal vectors to the specular MLP branch and to keep the specular MLP from being impacted by diffuse color, we exclusively train the diffuse MLP branch for a few training steps at the beginning, then jointly train both diffuse and specular branches.

%% file: method_light.tex
\vspace{-4pt}
\section{Decomposing Lighting and Reflectance}\label{sec:lighting}
In this section, we describe how we decompose the reflectance and environment light from the fully trained SDF-based PointNeRF model described in Sec. \ref{sec:sdf_pnerf}. There are three key components in our lighting estimation method. First, we use an HDR light probe image learned by a MLP to represent the environment light. Second, we employ the shadow mapping technique to approximate the surface light visibility. Third, we add extra MLP branches to estimate the surface BRDF properties. Our light estimation approach is designed assuming distant environment illumination and also does not consider indirect reflection. 
\subsection{Neural Implicit Environment Light\label{sec:hrdi}}
Surface visibility is a key factor in determining the rendering of hard cast shadows. However, environment illumination approximation methods such as spherical Gaussians (SG) \citep{zhang2021physg, boss2021nerd} cannot be integrated with surface visibility. Thus, in \model, we represent the environment light as an HDR light probe image (HDRI) \citep{debevec1998rendering}.
Every pixel on the light probe image can be treated as a distant light source on a sphere, thus the integral in the rendering equation (Eqn. \ref{eq:re}) can be discretized as the summation over a hemisphere of the light sources (pixels) on the light probe image. To reduce computational complexity, environment light is estimated at a $16 \times 32$ resolution in the latitude-longitude format, which can be approximately treated as 512 point light source locations on a sphere centered around the object.

Unlike NeRFactor \citep{nerfactor} which directly optimizes 512 pixels $\mathbf{p}_i$ as learnable parameters, we instead use a  coordinate-based MLP $\bm E$ to implicitly represent the light probe image. Given the $i$th light source's direction $\hat{\bm{\omega}}_i$ (normalized coordinate on a unit sphere), MLP $\bm E$ takes $\hat{\bm{\omega}}_i$ as input (with positional encoding \citep{mildenhall2020nerf}) to predict the corresponding light intensity $\bm E(\hat{\bm{\omega}}_i)$. 
This parameterization enables \model\ to learn more lighting details while maintaining spatial continuity \citep{tancik2020fourier}. 
Since our environment light model represents an HDR image, we adopt an exponential activation function \citep{mildenhall2022nerf} to convert the MLP output to the final light intensity.

\vspace{-5pt}
\subsection{Computing Visibility from Depth\label{sec:vis}}
Existing NeRF methods compute the surface light visibility via direct ray marching \citep{bi2020neural, nerfactor} or a visibility estimation MLP \citep{nerv2021, nerfactor}. Although these methods can make accurate visibility predictions, they are either compute-intensive or not applicable to geometry deformation. 
Therefore, we instead apply the shadow mapping technique \citep{williams1978casting} to approximate the light visibility. 
{The shadow mapping method determines the light visibility by checking the depth difference between the depth map from the current viewpoint and the depth maps from the light sources under the same coordinate system.}
We can utilize the estimated depth maps from our NeRF representation to indirectly calculate the light visibility of a point on the surface.
The visibilities of all light sources for a point $\mathbf{x}$ on the surface then form a visibility map $\bm{V}(\mathbf{x}, \hat{\bm{\omega}}_i)$ with the same shape as the light probe image. 

\subsection{Neural BRDF Estimation\label{sec:ref}}
Similar to other prior works \citep{nerv2021, boss2021neural}, we also use the microfacet BRDF \citep{walter2007microfacet}
to describe the surface reflectance property. We use three additional MLP branches, with interpolation point feature $\mathbf{f}_{\mathbf{x}}$ (Eqn. \ref{eq:interp_feat}) as input\footnote{\footnotesize The final prediction of BRDF parameters is the weighted sum of predictions of sampled ray points.}, to predict BRDF parameters (diffuse $\mathbf{a} \in \mathbb{R}^3$, specular $\mathbf{s} \in \mathbb{R}^3$, and roughness $\alpha \in \mathbb{R}$).
Recall that our NeRF model also decomposes radiance into diffuse color and specular color (Eqn. \ref{eq:d+s}), which is in accordance with the BRDF decomposition. Our trained diffuse and specular MLP ($R_d$ \& $R_s$) can be served as priors for learning the corresponding BRDF parameters. We show how  these priors are utilized as follows.

\textbf{Diffuse Color}. 
Lambertian (diffuse) color is mainly determined by the BRDF diffuse parameters $\mathbf{a}(\mathbf{x})$, also known as albedo. 
Since diffuse color can be perceptually treated as the shaded albedo, we directly initialize our albedo MLP branch $B_a(\mathbf{f}_{\mathbf{x}})$ with the weights from the diffuse MLP $R_d(\mathbf{f}_{\mathbf{x}})$. This provides a reasonable starting point for learning the real albedo.


\textbf{Specular Color}. We use another two MLP branches, $B_s(\mathbf{f}_{\mathbf{x}})$ and $B_r(\mathbf{f}_{\mathbf{x}})$, to model the specular BRDF parameters specular $\mathbf s$ (also known as Fresnel F0) and roughness $\alpha$. Unlike the albedo branch $B_s(\mathbf{f}_{\mathbf{x}})$,
we cannot initialize these specular BRDF branches with our specular radiance MLP $R_s$, as $R_s$ is conditioned on view directions and normals. Thus we need to optimize $B_s$ and $B_r$ from scratch. In the meanwhile,
the results from the specular MLP $R_s$ can be used as one loss term to supervise the learning of specular BRDF. 

\subsection{Training}\label{sec:training_s2}
\begin{figure}
    \vspace{-6pt}
    \centering
    \includegraphics[width=\linewidth, trim={0pt 0pt 0 125pt}, clip]
    {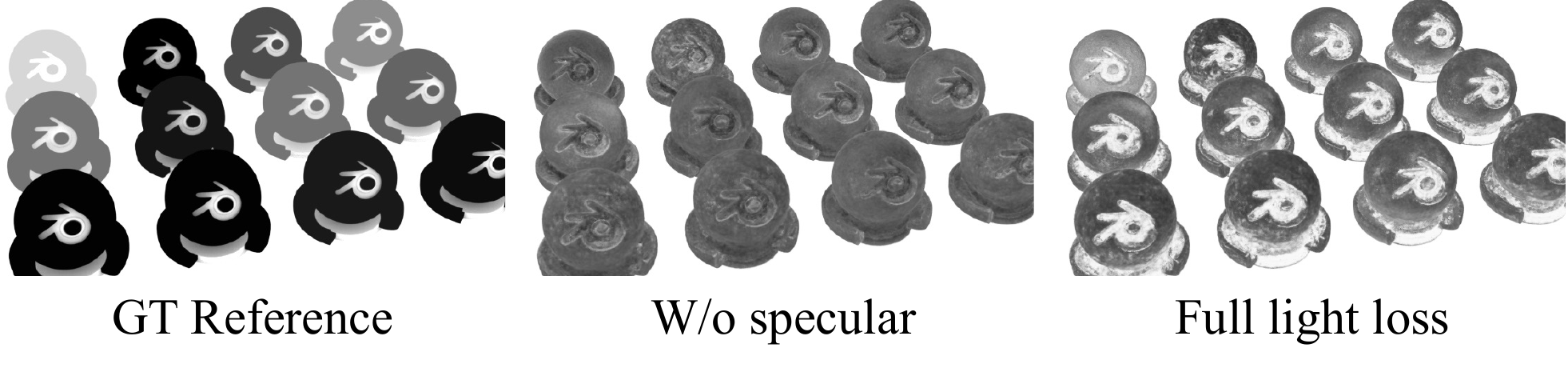}
    \vspace{-20pt}
    \caption{The effects of the specular color loss on the estimated material roughness. The additional specular color loss can help \model\ better distinguish different materials in the scene.}
    \label{fig:roughness}
    \vspace{-18pt}
\end{figure}

During the training of lighting estimation, we jointly optimize environment light and BRDF parameters.
In addition to using ground truth color $\mathbf{C}^*$ for supervision, we also use specular color $\mathbf{C}_s$ obtained from MLP $R_s$, to supervise the BRDF integrated specular color $\mathbf{C}_s^\prime$. This additional supervision is helpful for \model\ to learn more accurate reflectance BRDF parameters (see Fig. \ref{fig:roughness} and supplement).
Because we use precomputed depth for visibility map during this training step, we freeze the first part of our model that outputs predicted SDF values $d(\mathbf{x})$ and radiance features $\mathbf{f}_{\mathbf{x}}$, and only optimize these BRDF MLP branches ($B_d$, $B_s$ and $B_r$) and our environment light model $\bm E$. The loss function is formulated as 
\vspace{-6pt}
{\small
\begin{align}
    \mathcal{L}_{light} =& \lambda_{c} \mathcal{L}_{color}(\mathbf{C}^*, \mathbf{C}_d^\prime + \mathbf{C}_s^\prime) + \lambda_{g} \mathcal{L}_{color}(\mathbf{C}_s, \mathbf{C}_s^\prime) 
    \label{eq:loss_light}
\vspace{-9pt}
\end{align}
}
where $\lambda_{c}$
and $\lambda_g$ are loss weight hyperparameters. Since we already have strong geometry and radiance priors from the trained NeRF model, 10k iterations of training in this step are sufficient for model convergence. 

%% file: method_deform.tex
\section{Geometry Editing}\label{sec:deformation}
As described in Section \ref{sec:sdf_pnerf}, our hybrid NeRF model employs the point cloud as the explicit representation.
Neural points in our final optimized neural point cloud align well with the object surface.
This enables users to select desired parts easily and perform common geometry transformations, including translation, rotation, and scaling. The local topology and neural point features are invariant to such transformations, thus we can render the deformed scene without retraining or updating any neural parameter. Figure \ref{fig:demo} shows an example of direct manipulation.

Since our SDF-based scene representation captures higher quality surfaces, we can extract the object mesh from our neural representation (i.e., marching cube \citep{marching_cube} on the zero-level set of predicted SDF).
The mesh then acts as a proxy to guide the point movements in the point cloud.
Users can apply existing mesh deformation methods (e.g., ARAP \citep{sorkine2007rigid} and Cage \citep{ju2005mean}) to the extracted mesh.
The mesh's deformation is then transferred to the point cloud. 
We first register each neural point to its closest triangle face on the mesh, then compute the corresponding projected barycentric coordinates with signed distances. Based on the deformed mesh, we apply a barycentric interpolation to find the new projected point locations and normals, then unproject the points along the normals by the signed distances.  

{
As we provide two ways to perform geometry editing, users can choose the one that fits them best. Direct point cloud manipulation is better at rigid-body movements for mechanic components (e.g., cars and toys). While mesh-guided deformation is better at non-rigid body deformations (e.g., more natural pose deformation for human body).}

%% file: experiments.tex
\vspace{0pt}
\section{Experiments}
\vspace{-4pt}
We evaluate our method on various challenging 3D scenes. We make comparisons with prior works based on the quality of view synthesis, lighting estimation, and geometry deformation.

\noindent
\textbf{Datasets.}
We use 2 two datasets for evaluation: 8 NeRF's Blender synthetic scenes \citep{mildenhall2020nerf} and 4 real-captured scenes in BlendedMVS \citep{yao2020blendedmvs} that are processed by NSVF \citep{liu2020neural}. 
To quantitatively evaluate the rendering quality of deformed scenes, we include two additional Blender scenes, Mannequin (CC-0) and T-Rex (CC-BY), which contain the rigged meshes for providing precise references for the scene deformations.

For clarity, in the following experiments, \emph{SPIDR} refers to our PBR rendering (with estimated BRDF), while \emph{SPIDR$^*$} refers to our non-PBR volume rendering.

\subsection{View Synthesis and Reconstruction}
\label{sec:exp_nvs}

\begin{figure}[t]
    \centering
    \vspace{-13pt}
    \includegraphics[width=\linewidth,trim={35pt 0 0 5pt},clip]{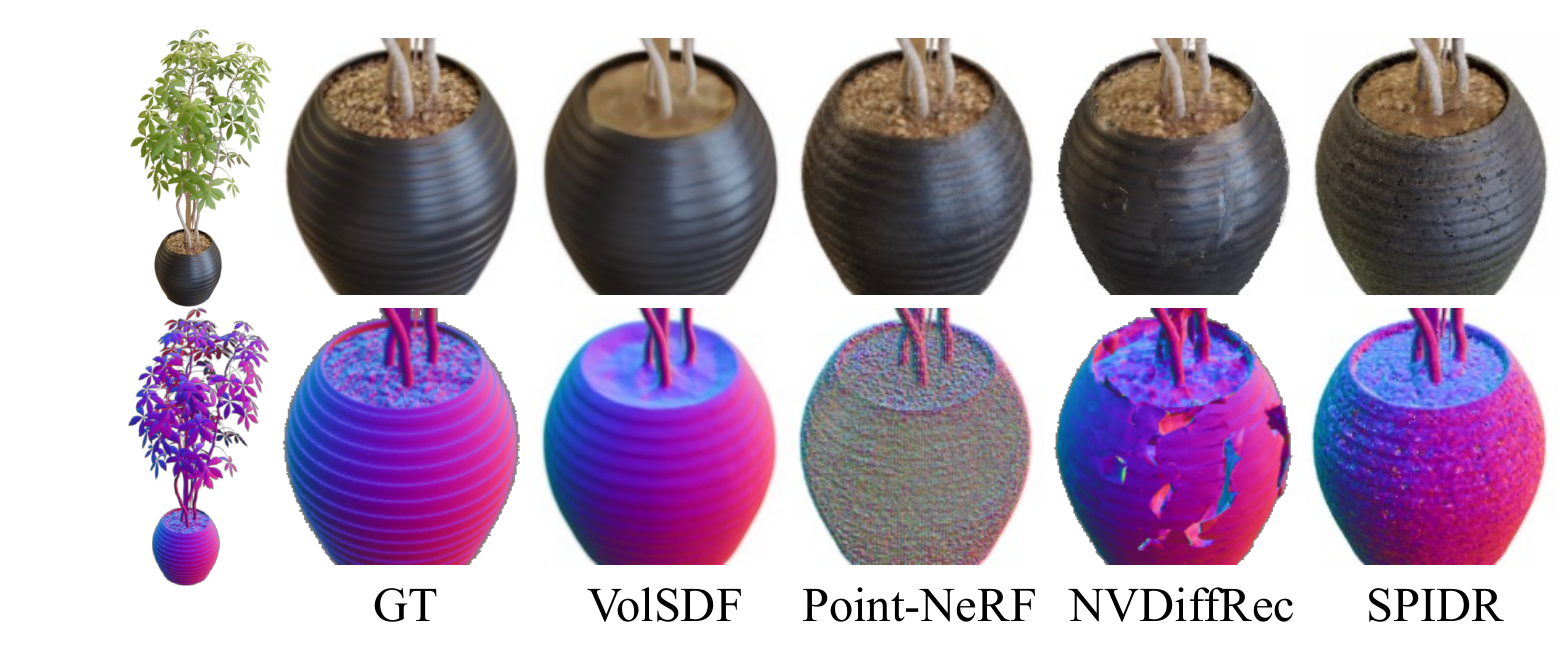}
    \vspace*{-21pt}
    \caption{Rendering results and estimated normals for Ficus scene.}
    \label{fig:nvs}
    \vspace{-14pt}
\end{figure}

We first evaluate the rendering quality of our non-PBR model (SPIDR$^*$) for static scenes using PSNR, SSIM, and LPIPS \citep{zhang2018unreasonable}. We also evaluate the quality of estimated surface normals with mean angular error (MAE). Results are summarized in Table \ref{tab:render_scores} and illustrated in Figure \ref{fig:nvs}.
%

\begin{table}[t]
    \footnotesize
    \centering
    \vspace{-8pt}
    \begin{tabular}{ll|rrrr}
    \hline
    \multicolumn{2}{c|}{ } & \textbf{PSNR}$\uparrow$ & \textbf{SSIM}$\uparrow$ & \textbf{LPIPS}$\downarrow$ & \textbf{MAE}\si{\degree}$\downarrow$ \\
    \hline
    \multirow{5}{*}{\rotatebox[origin=c]{90}{\textbf{Synthetic}}}
    & VolSDF & 27.96 & 0.932 & 0.096 & \textbf{19.87} \\
    & NVDIFFREC & 29.05 & 0.939 & 0.081 & \underline{25.01} \\
    & PointNeRF & \textbf{33.31} & \textbf{0.978} & \textbf{0.027} &     48.09 \\
    & SPIDR$^*$ & \underline{32.30} & \underline{0.976} & \underline{0.029} & 30.28\\
    & SPIDR & 27.78 & 0.951 & 0.061 & 27.08  \\ 
    \hline
    \multirow{5}{*}{\rotatebox[origin=l]{90}{\textbf{B.MVS}}}
    & VolSDF & 24.77 & 0.915 & 0.148 & \textbf{25.95} \\
    & PointNeRF & \underline{26.71} & \underline{0.941}  & \underline{0.078} & 52.10 \\
    & SPIDR$^*$ & 26.65 & \textbf{0.950} & \textbf{0.063} & 47.48 \\
    & SPIDR & 24.71 & 0.921 & 0.098 &  \underline{34.78}  \\ 
    \hline
    \end{tabular}
    \vspace{-5pt}
    \caption{Quantitative comparison with baseline methods. Note that in the MAE column, \emph{SPIDR$^*$} indicates computed gradient normal, while \emph{SPIDR} indicates point interpolated normal.\label{tab:render_scores}}
\end{table}

Our results demonstrate that SPIDR$^*$ generates rendering results that are comparable to the original Point-NeRF and superior to NVDIFFREC \cite{munkberg2021extracting} and VolSDF \cite{yariv2021volume}. Additionally, our method provides high-quality normal estimation, significantly better than Point-NeRF. The interpolated point normal shown in \model\ further filters out high-frequency noise in computed gradient normal and provides a smoother normal representation for our PBR rendering. 


In comparison to SPIDR$^*$ and Point-NeRF, our PBR model \model\ exhibits lower PSNR values. This is due to PBR's reliance on surface quality, where even a slight deviation in the predicted normal can lead to noticeable rendering artifacts. Furthermore, \model's simplified rendering equation cannot accurately capture high-frequency, view-dependent specular reflections. As a result, \model\ struggles in glossy or shiny scenes, such as those depicted in Mic, Materials, and Drums. Although complex glossy reflections were not a focus of our research, we acknowledge opportunities for future improvement. Despite this, our PBR rendering still performs comparably to VolSDF.

\begin{figure}[t]
    \vspace{-15pt}
    \centering
    \small
    \includegraphics[width=\linewidth]{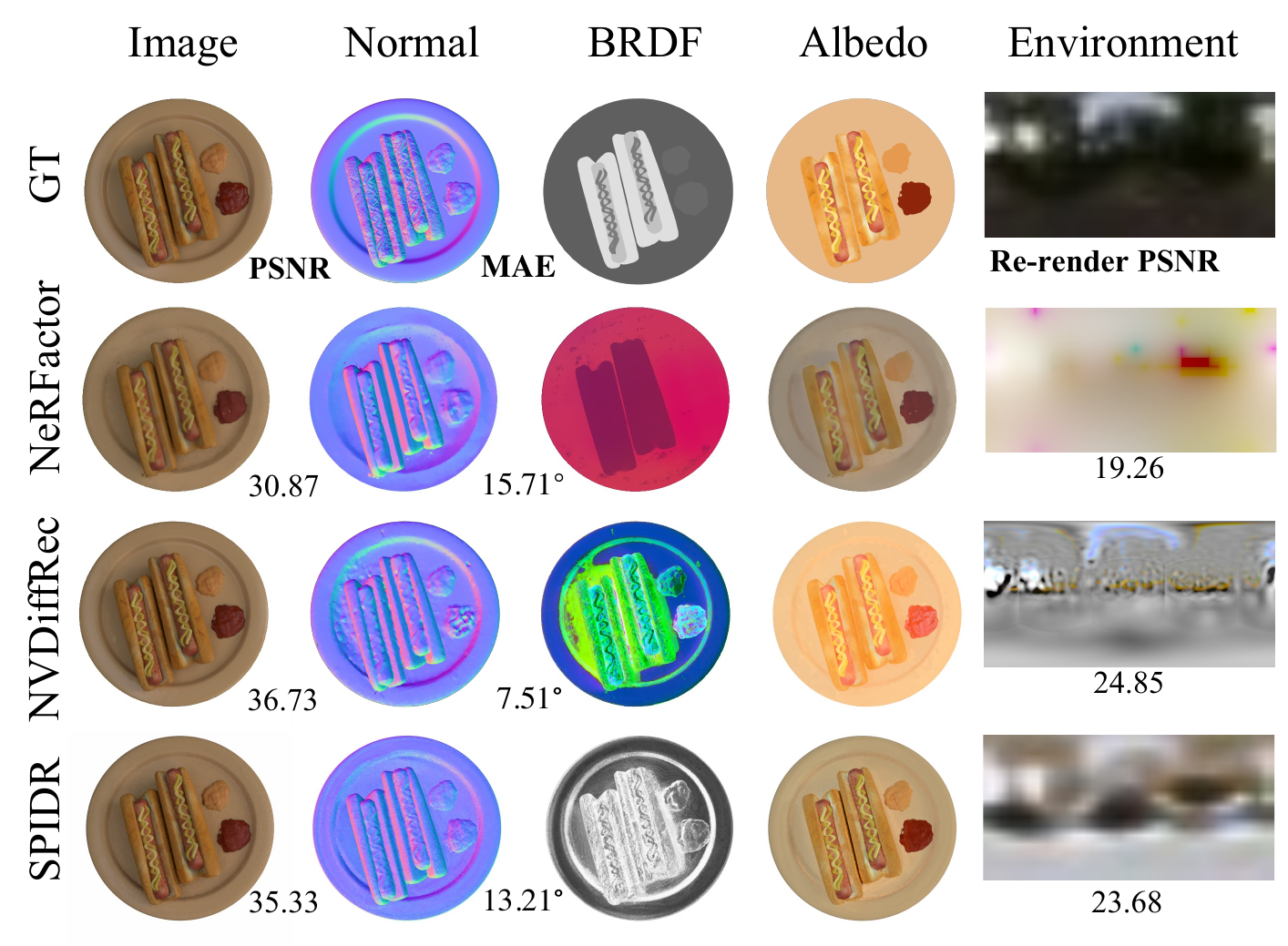}
    \vspace{-15pt}
    \caption{Lighting and BRDF estimation. \label{fig:light_compare} In the BRDF column, NeRFactor visualizes the learned BRDF latent code. NVDIFFREC shows the roughness/metalness texture. \model\ shows the grayscale roughness. HDR light probes are converted to sRGB.
    }
    \vspace{-15pt}
\end{figure}

\subsection{Inverse Rendering}\label{exp:light}
To demonstrate the quality of estimated illumination parameters such as environment light and material properties, we compare against other neural inverse rendering methods \cite{nerfactor, munkberg2021extracting}, using NeRF synthetic scenes provided by NeRFactor. Figure \ref{fig:light_compare} shows the results for the Hotdog scene. 
We observe \model\ has PSNR and MAE scores on par with NVDIFFREC and significantly better than NeRFactor. 
To evaluate the environment estimation, we use Blender to re-render scenes with estimated light probes. We then measure the PSNR between re-rendered images and images with the ground truth environment. 
\model\ can faithfully recover 3 light sources shown in the ground truth environment.
We use scene relighting to indirectly evaluate the estimated BDRF. The results in Table \ref{tab:nerfactor_spidr} demonstrate that \model\ has a competitive performance on light decomposition and NeRF-based inverse rendering.

\begin{table}[h]
\vspace{-5pt}
\input{tables/compare_with_nerfactor.tex}
\vspace{-5pt}
\caption{Quantitative results (PSNR/SSIM) of scene relighting. 
The scores are averages across 8 light probes included in Blender, with each light probe containing 100 sampled views.
\label{tab:nerfactor_spidr}
}
\vspace{-6pt}
\end{table}



\vspace{-5pt}
\subsection{Geometry Editing}
\label{sec:exp_geo}

\begin{table}[t]
    \vspace{-4pt}
    \centering
    \begin{tabular}{l|c|c}
    \hline
        &  \textbf{Manikin} (7 Poses) & \textbf{T-Rex} (2 Posese) \\ \hline
        Xu \etal \cite{xu2022deforming} & 24.32/0.951/0.060 & 24.06/0.941/0.037 \\
        NVDiffRec & 28.30/0.972/0.024 & 26.67/0.949/0.044\\
        \model$^*$ & 31.39/0.987/0.016 & 33.42/\textbf{0.989/0.010} \\
        \model & \textbf{32.93/0.990/0.015} & \textbf{33.47}/0.987/0.013 \\
        \hline
    \end{tabular}
    \vspace{-5pt}
    \caption{Quantitative evaluation (PSNR/SSIM/LPIPS)  of rendering for deformed scenes. The scores are averages across multiple deformed poses, with each pose containing 20 sampled views.}
    \label{tab:deform_score}
    \vspace{-2pt}
\end{table}

To demonstrate the effectiveness of \model in rendering deformed scenes, we compare our model against deforming-NeRF \cite{xu2022deforming} and NVDIFFREC \cite{munkberg2021extracting}.  
To obtain quantitative metrics, we directly deform two rigged Blender scenes (Mannequin and T-Rex) and use ground truth meshes or bones to guide the deformations of the evaluated models. 
Our results, presented in Table \ref{tab:deform_score} and Figure \ref{fig:deform_compare}, show that \model\ outperforms the baseline models, including our non-PBR model \model$^*$. This is due to \model's ability to update illumination based on geometry occlusions, resulting in more accurate image synthesis for deformed scenes, as highlighted in Figure \ref{fig:deform_compare}. 
Notably, \model\ is the only evaluated neural rendering model that achieves adaptive illumination updates. While NVDIFFREC can also render updated illuminations, it employs the existing Blender path tracer renderer, which is not a typical neural rendering approach, and our rendering quality surpasses that of NVDIFFREC.
Moreover, our point-based editing method allows for more precise geometry editing than other approaches: Cage-based methods such as deforming-NeRF \cite{xu2022deforming} may struggle in complex deformations; NVDIFFREC's mesh deformation is limited by the mesh topology.

\begin{figure}[t]
    \vspace{-5pt}
    \centering
    \includegraphics[width=\linewidth]{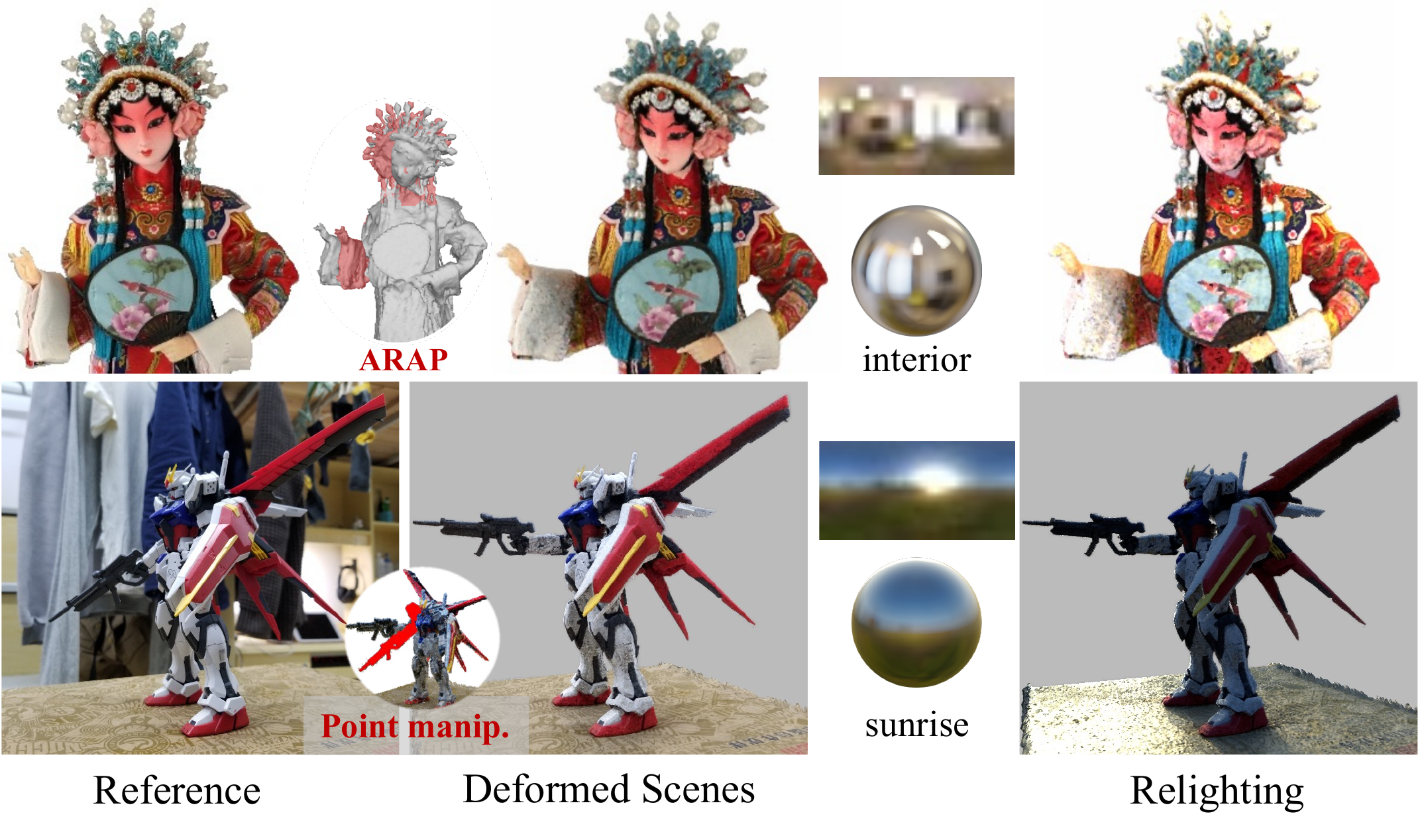}
    \vspace{-20pt}
    \caption{Additional results to showcase SPIDR's rendering capability in deformation and illumination.}
    \label{fig:add_res}
    \vspace{-15pt}
\end{figure}

In addition to scenes used for comparison, we also use \model\ to render and relight other challenging scenes with various deformations, as depicted in Figure \ref{fig:add_res}. Please refer to the supplementary materials for more results.

\subsection{\ruofan{Ablation Study}}
\label{sec:ablation}
\textbf{Ablation on SDF regularizations.}
We plot the SDF and compositing weight curves along the sampled ray points to show the effects of our proposed SDF regularization in Figure \ref{fig:ablation_sdf}. 
Compared to models without the SDF continuity regularization $\mathcal{L}_d$ (Eqn. \ref{eq:l_d}) or negative sparsity regularization $\mathcal{L}_s$ (Eqn. \ref{eq:l_s}), 
the model with both $\mathcal{L}_s$ and $\mathcal{L}_d$ (``w/ $\mathcal{L}_s$, w/ $\mathcal{L}_d$") has more stable SDF decreasing curve when the sampled points cross the object surface (no oscillation or horizontal line). 
Consequently, the corresponding compositing weight curve with one concentrated spike also indicates more accurate and smooth geometry estimations.

\begin{figure}[!t]
    \centering
    \vspace*{-8pt}
    \includegraphics[width = \linewidth, trim={0 0 5pt 0},clip]{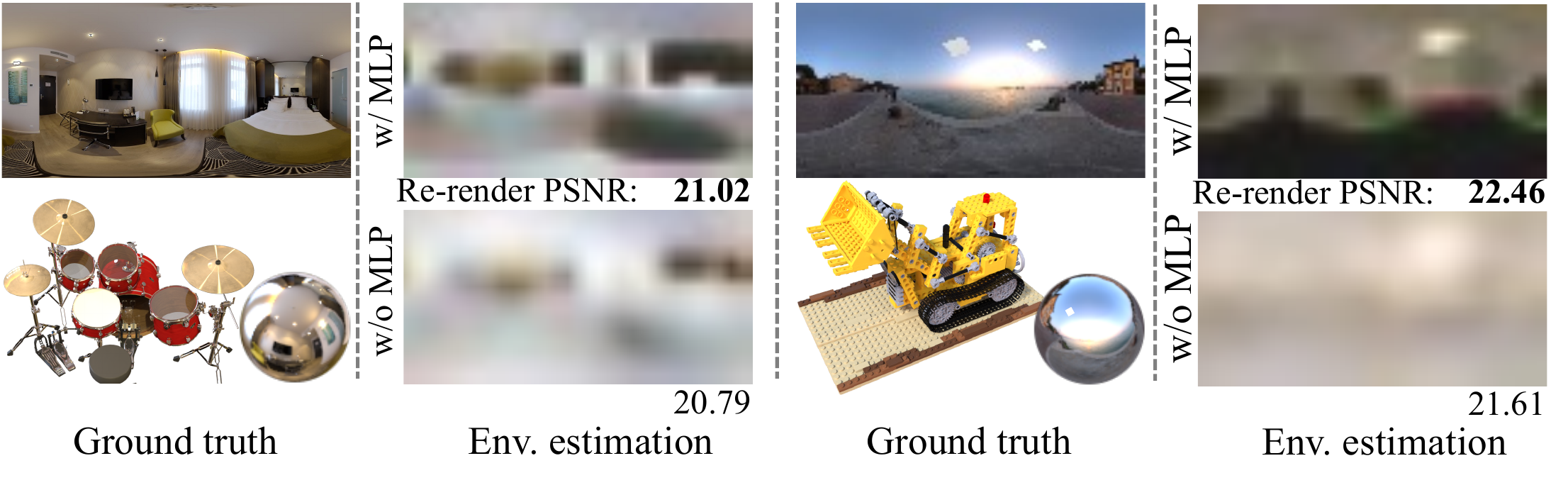}
    \vspace*{-20pt}
    \caption{The comparison of environment light estimation. Note that the original Lego Blender scene has a much brighter external light source (bright spots on the estimation) than the used HDRI.
    \label{fig:env_compare}}
    \vspace{-5pt}
\end{figure}
\begin{figure}[t]
    \vspace{-3pt}
    \includegraphics[width = \linewidth,trim={28pt 0 10pt 43pt},clip]{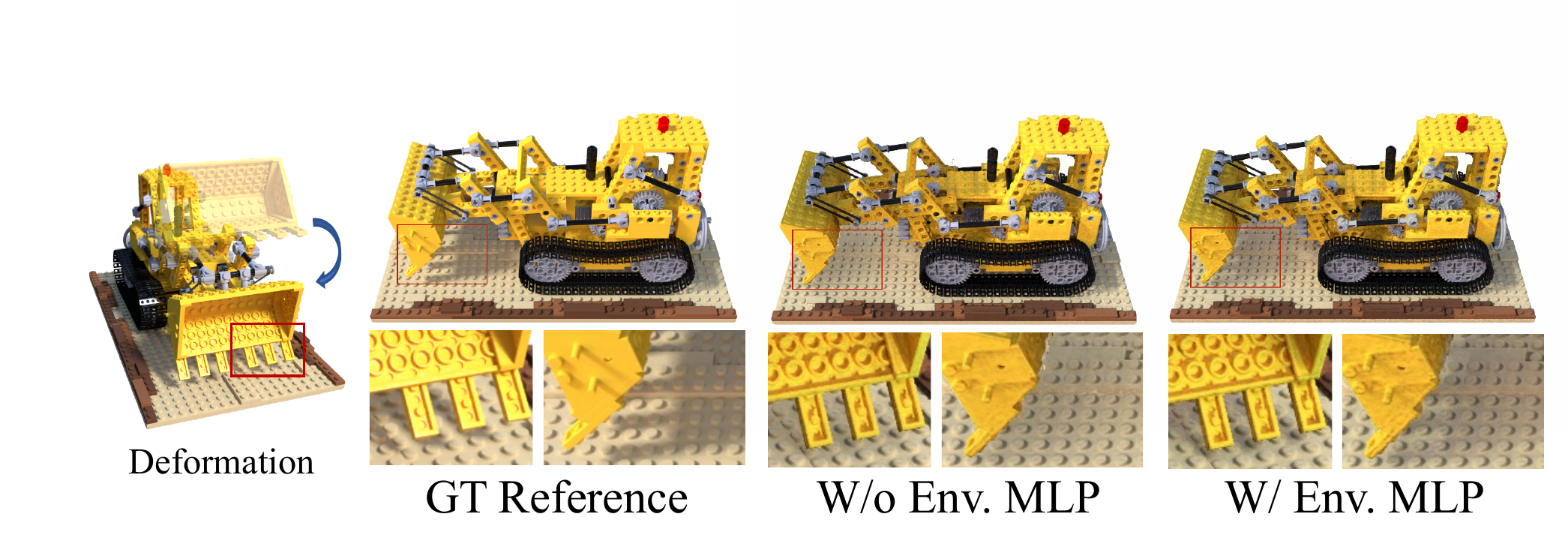}
    \label{fig:lego_def_w_mlp}
    \vspace*{-20pt}
    \caption{\model's PBR rendering results of the deformed Lego scene using estimated environment shown in Fig \ref{fig:env_compare}.}
    \label{fig:illumination_deform}
    \vspace{-17pt}
\end{figure}

\textbf{Ablation on environment light representations}
To show the effectiveness of our MLP-based environment light representation. We compare our MLP-based representation with a non-MLP light representation, which directly learns light probe pixels \citep{nerfactor}. As shown in Figure \ref{fig:env_compare}, our MLP representation is able to recover more  details in the estimated environment light (e.g., windows in Drums' environment) compared to non-MLP representation.
The high-intensity and concentrated light sources in the light probe image are critical for synthesizing hard-cast shadows. As demonstrated in Figure \ref{fig:illumination_deform}, the over-smoothed light probe estimated by the non-MLP representation cannot synthesize shadows for the deformed Lego as clearly as the results from our MLP-based environment light representation.


%% file: tables/compare_with_nerfactor.tex
\centering
\begin{tabular}{l|c|c|c}
\hline
          & \textbf{Hotdog} & \textbf{Lego} & \textbf{Avg.} \\
\hline
NeRFactor & 25.96/0.940   & \textbf{26.03}/0.913 & 26.00/0.927 \\
NVDiffRec & \textbf{28.77}/0.932       &   21.03/0.846   &  24.90/0.889    \\
SPIDR     & 27.83/\textbf{0.946}       &   \textbf{26.03}/\textbf{0.938}   &  \textbf{26.93}/\textbf{0.932} \\
\hline
\end{tabular}

%% file: conclusions.tex
\vspace{-6pt}
\section{Conclusion}
This paper presents \model, a novel hybrid SDF-based 3D neural representation that enables the more accurate rendering of scene deformations and illumination. Our proposed neural representation, augmented with new SDF regularizations, enables superior reconstruction of scene surfaces. Additionally, our neural representation of the light probe allows \model\ to capture more precise environment highlights, resulting in better rendering of shadowing effects. These contributions make \model\ the first approach capable of rendering scenes with fine-grained geometry deformations while producing more accurate illumination.

\textbf{Limitations}.
 As discussed in \ref{sec:exp_nvs}, the PBR rendering for scenes sacrifices some rendering quality. 
 In inverse rendering,  high rendering quality is often contradictory to accurate illumination decomposition because shadows and lighting can be easily misinterpreted  as textures. 
 Thus, rendering quality still requires joint efforts in surface reconstruction and inverse rendering with room for future work. 
 Additional post-processing such as image denoising \citep{ulyanov2018deep} could also enhance the rendering quality.
Second, \model\ cannot completely remove existing shadows when trained on a single unknown environment. One possible solution is to use images from multiple environments \citep{boss2021nerd, boss2021neural}).
Additionally,  \model\ currently only considers direct illumination, which means that it cannot capture shading effects caused by indirect illumination. This is another avenue for future research.


%% file: appendix.tex
\appendix

\input{appendices/light_vis.tex}

\input{appendices/deformation.tex}

\input{appendices/impl_details.tex}

\input{appendices/point_grow_prune.tex}

\input{appendices/additional_eval.tex}

\input{appendices/ablation.tex}

\input{appendices/additional_res.tex}

%% file: appendices/light_vis.tex
\section{Details of Physically-based Rendering}
\subsection{Light Visibility}
In Section~\ref{sec:vis}, we briefly demonstrate how we compute visibility maps from depth via the shadow mapping technique. We give more details in this section.
Shadow mapping, a common method in computer graphics \cites{williams1978casting, eisemann2013efficient}, aims to recover the per-point visibility maps from shadow maps of all light sources. 
The shadow maps are computed by comparing the depth difference between the depth map (z-buffer) from the view direction and the depth maps from the light sources after projecting to the same coordinate system. 

Similar to other NeRF models, we use the expected termination point of each camera ray from NeRF's ray marching to approximate the depth map $\bm{D}$. 
Given the depth map $\bm{D}_c$ of the image captured by a camera with pose $\bm{T}_c$, and the depth maps $\bm{D}_{\bm{\omega}_i}$ of light sources with pose $\bm{T}_{\bm{\omega}_i}$, a pixel $(u, v)$ that represents a surface point $\mathbf{x}$ from the camera $\bm{T}_c$ has the shadow $\bm{S}(u, v, \hat{\bm{\omega}}_i)$ for the $i$th light source:
\begin{equation}
    \bm{S}(u, v, \hat{\bm{\omega}}_i) = \mathbf{1}\left\{\text{interp}(\bm{D}_{\bm{\omega}_i}^{\prime}, u,v) - \bm{D}_c(u,v) + b\right\}
\end{equation}
where $\mathbf{1}\{\cdot\}$ is a zero-one indicator function, $\bm{D}_{\bm{\omega}_i}^{\prime} = \bm{T}_{c}\bm{T}_{\bm{\omega}_i}^{-1}\bm{D}_{\bm{\omega}_i}$, representing the depth maps after projecting to $\bm{T}_c$'s coordinate system, $b$ is the shadow bias term, $\text{interp}(\cdot)$ represents a bi-linear interpolation function. The visibility map $\bm{V}$ is a collection of values from shadow maps of all light sources $\bm{V}(\mathbf{x}, \hat{\bm{\omega}}_i) =\bm{S}(u, v, \hat{\bm{\omega}}_i)$. 
Since we treat our environment light as 512 light sources, 512 depth maps of the scene from each light source need to be pre-computed. To minimize this compute overhead, we render half-resolution light depth maps, which can be finished within 10 minutes for our tested scenes, but still provides reasonably accurate visibility maps from our observations.

\subsection{The Approximation of the Rendering Equation }
Given light probe image $\bm E$ and visibility map $\bm V$, we can approximate the discretized incoming radiance $L_i$ with the product of light probe and its visibility, that is $L_i(\mathbf{x},  \hat{\bm{\omega}}_i) =  \bm E(\hat{\bm{\omega}}_i) \bm V(\mathbf{x},  \hat{\bm{\omega}}_i)$. The rendering equation (Eqn. \ref{eq:re}) can thus be approximated as an accumulation of the irradiance from all discretized light directions:
\begin{align}
        L_o(\mathbf{x}, \hat{\bm{\omega}}_o) &\approx \sum_{\hat{\bm{\omega}}_i}  L_i(\mathbf{x},  \hat{\bm{\omega}}_i) B(\mathbf{x}, \hat{\bm{\omega}}_i, \hat{\bm{\omega}}_o) (\hat{\mathbf{n}}\cdot \hat{\bm{\omega}}_i)  \Delta \hat{\bm{\omega}}_i \notag \\
        &=\sum_{\hat{\bm{\omega}}_i}  E(\hat{\bm{\omega}}_i) V(\mathbf{x},  \hat{\bm{\omega}}_i) B(\mathbf{x}, \hat{\bm{\omega}}_i, \hat{\bm{\omega}}_o) (\hat{\mathbf{n}}\cdot \hat{\bm{\omega}}_i)  \Delta \hat{\bm{\omega}}_i 
        \label{eq:re_sum}
\end{align}
where $\Delta \hat{\bm{\omega}}_i$ denotes the solid angle for the corresponding incoming light direction $\hat{\bm{\omega}}_i$ from the light probe image's hemisphere.

%% file: appendices/deformation.tex
\section{Mesh-Guided Point Cloud Deformation}
In this section,  we give more details on how we perform the mesh-guided point cloud manipulation discussed in Sec. \ref{sec:deformation} of the main paper. 
Extracting the guidance mesh from our neural representation is the first step for mesh-guided deformation. There are two approaches we can use to extract the mesh. The first approach is to perform the Marching Cube algorithm \cites{marching_cube} to convert the zero-level set of the SDF estimated by our model into the triangle mesh.
The second approach is to directly convert our model's oriented point cloud representation to the mesh using point cloud-based mesh reconstruction methods such as Screened Poisson algorithm \cites{screened}.
We interchangeably use both approaches to extract meshes. The Marching Cube method usually captures more accurate object surfaces, while the point cloud method can maintain more thin structures in the object.
Given the extracted object mesh, we can perform any mesh-based deformation method to edit the object geometry. We mainly use ARAP (as-rigid-as-possible) method \cites{sorkine2007rigid} to perform the mesh deformations in this paper.

After obtaining the deformed mesh, the next step is to transfer the deformations made on the guidance mesh to the point cloud.
The detailed procedure is described in Algorithm \ref{algo:meshdeform}. 
Specifically, we project each point in the point cloud onto its closest triangle face in the mesh and calculate the corresponding barycentric coordinate \cites{ericson2004real} and point-to-face signed distance (Line 3-5 in Algorithm \ref{algo:meshdeform}). 
These projected points on the triangle faces are moved in correspondence with the triangle faces. The positions of projected points after the deformation are determined by the exact barycentric coordinates calculated before (Line 7-8 in Algorithm \ref{algo:meshdeform}).
We then offset those projected points along the normals of deformed faces by the precomputed signed distances (Line 10 in Algorithm \ref{algo:meshdeform}). Finally, these offset points form a new point cloud representing the deformed object.

{
\IncMargin{1.2em}
\begin{algorithm}[h]
\caption{Mesh Guided Deformation\label{algo:meshdeform}}
\SetKwInOut{Input}{Input}
\SetKwInOut{Output}{output}
\SetKwComment{Comment}{/* }{ */}
\SetKwInOut{Define}{Functions}
\Input{neural point cloud $\mathcal{P}$, extracted mesh ($\mathcal V, \mathcal F$), deformed mesh ($\mathcal V^\prime, \mathcal F^\prime$)}
$\mathcal{P}^\prime \leftarrow \{\}$;\\
\For{$P_i$ in $\mathcal{P}$}{
    $F_i \leftarrow$ the nearest face to $P_i$;\\
    $\bar P_i \leftarrow$ the point projected from $P_i$ onto $F_i$;\\
    $d_i \leftarrow $ signed distance for $P_i$ to $F_i$; \\
    $\alpha_i, \beta_i, \gamma_i \leftarrow $ barycentric coordinate of $\bar P_i$; \\ 
    $V_{0,i}^\prime, V_{1,i}^\prime, V_{2,i}^\prime \leftarrow $ three corners of $F_i^\prime \in \mathcal F^\prime$; \\
    $\bar P_i^\prime \leftarrow \alpha_i V_{0,i}^\prime+\beta_i V_{1,i}^\prime+\gamma_i V_{2,i}^\prime$;\\
    $\hat{\mathbf{n}}^\prime_{i} \leftarrow \text{triangle interpolated normal of } \bar P_i^\prime$;\\
    $\mathcal{P}^\prime \leftarrow \mathcal{P}^\prime \cup \{\bar P_i^\prime + d_i\hat{\mathbf{n}}^\prime_{i}\}$;
}
return $\mathcal{P}^\prime$;
\end{algorithm}
\DecMargin{1.2em}
}

%% file: appendices/impl_details.tex
\section{Implementation Details}
\begin{figure*}[hbtp]
    \centering
    \small
    \vspace{-10pt}
    \includegraphics[width=0.8\textwidth,trim={0cm 0 0cm 0},clip]{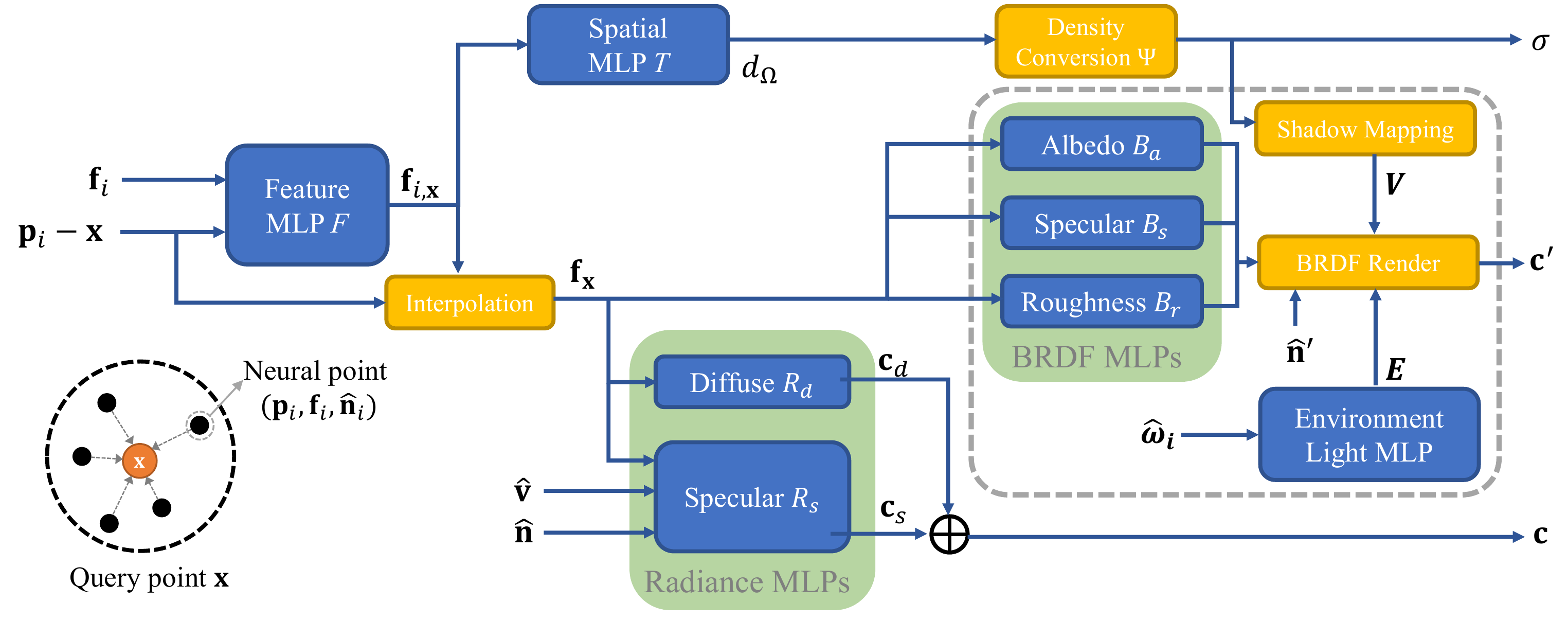}
    \vspace*{-2pt}
    \caption{
     A visualization of \model's architecture. 
     For simplicity, we omit the computation flow for both SDF computed normal $\hat{\mathbf{n}}$ and interpolated point normal $\hat{\mathbf{n}}$ (Eqn. \ref{eq:interp_normal}).
    }
    \label{fig:arch}
    \vspace{-15pt}
\end{figure*}

\textbf{Architecture and hyperparameters}.
We describe how our point-based neural implicit model is designed in this section.
Figure \ref{fig:arch} shows the overview of our model design. The feature and spatial MLP have 4 layers and 1 layer respectively. The conditioned neural feature $\mathbf{f}_{i, \mathbf{x}}$ has a feature size of 256. The radiance and BRDF MLP branches are 3-layer MLPs\footnote{We use 5-layer specular radiance MLP for the Materials scene, as it contains more challenging metallic reflections.} with 128 hidden units. 
Our environment light MLP is a simple 3-layer MLP with 128 hidden units. 
The relative position vector $\mathbf{p}_i - \mathbf{x}$, view direction $\hat{\mathbf{v}}$, and incoming light direction $\hat{\bm{\omega}}_i$ are encoded by the positional encoding (PE) \cites{mildenhall2020nerf} (with PE level of 5, 4, 3, respectively), before feeding into MLPs. In addition, \model's final accumulated pixel colors are all converted to the sRGB by using a linear-to-sRGB tone mapping function \cites{anderson1996proposal}.
Since MVS and point initialization step are not the main topics of this work, \model\ remains the same model design as Point-NeRF \cites{xu2022point} in the neural point cloud initialization step. In all our experiments, we use an MVSNet-based approach \cites{yao2018mvsnet} to get the initial point cloud by default.

\textbf{Training details.} After obtaining the initial neural point cloud via MVS approaches,
we start training \model\ on a single Nvidia RTX3090 GPU with Adam optimizer \cites{kingma2014adam} for both two training stages discussed in the main paper. 
We sample approximately 2048 rays per iteration for the training in both stages. Similar to \cites{liu2020neural, xu2022point}, we utilize the voxelized explicit representation to efficiently skip sampling points in the empty space.
In the non-BRDF NeRF training stage (Sec. \ref{sec:train_s1}), we train our model for 160k iterations. We empirically set the loss hyperparameters $\lambda_n=1e-3, \lambda_s=2e-3, \lambda_d=5e-3$ ($\lambda_s$ also varies depending on the specularity of the scene).  We set the initial learning rate to 5e-4 for MLPs and 2e-3 for neural point features. In order to get relatively stable normals for the specular MLP branch $R_s$, we skip the training of $R_s$ branch and treat the diffuse radiance as the final radiance during the first 10k training iterations.
In the BRDF training stage (Sec. \ref{sec:training_s2}),  we train the BRDF modules (BRDF MLP branches and environment light model) and environment light MLP for 10k iterations. We set the loss hyperparameters $\lambda_c = 1.0, \lambda_g = 1.0$. The initial learning rates for BRDF MLP branches and the environment light MLP are set to 5e-4  and 1e-3, respectively.

\textbf{Runtime.} The first stage of NeRF training with 160k iterations takes about 7 hours on one RTX3090. The second stage of BRDF training with 10k iterations only requires 0.5 hours. The precompute of depth maps with a halved resolution for shadow mapping only takes 5 minutes. 
In terms of inference speed, rendering one $800\times800$ image takes roughly 3.5 seconds and 4 seconds for non-PBR and PBR rendering, respectively. Since our code base is not optimized for runtime performance, further speedup can be achieved by various approaches such as early ray termination, customized CUDA kernels, and mixing precision computing.

%% file: appendices/point_grow_prune.tex
\section{Point Pruning and Growing}
\begin{figure*}[t]
    \centering
    \begin{subfigure}{0.292\textwidth}
    \includegraphics[width = \linewidth]{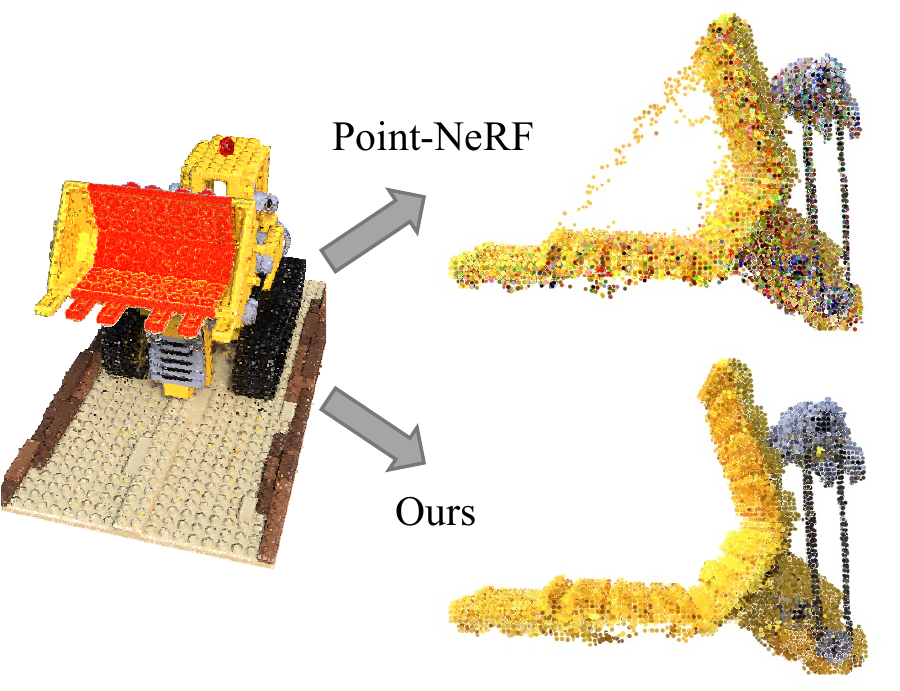}
    \vspace{-8pt}
    \caption{Point cloud after training}
    \label{fig:pcd_compare}
    \end{subfigure}
    \begin{subfigure}{0.7\textwidth}
    \includegraphics[width = \linewidth]{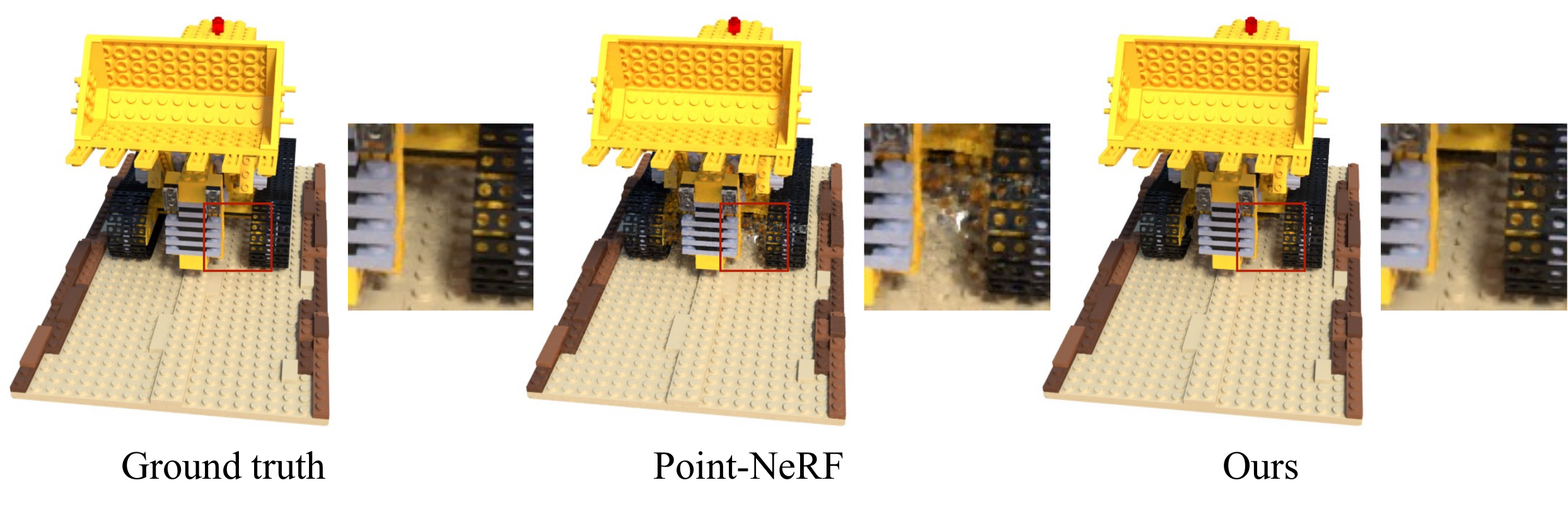}
    \caption{Final rendering results on the region where the initial points are sparse.}
    \label{fig:growing_compare}
    \end{subfigure}
    \caption{The comparison of point cloud pruning/growing on the Lego scene. Point-NeRF and our model have the same point cloud initialization before the NeRF training.}
    \label{fig:growing_pruning}
    \vspace{-8pt}
\end{figure*}

Although the original Point-NeRF \cites{xu2022point} has its own point pruning/growing mechanisms and performs well in various scenes, it still has limitations that can impact the rendering quality. 
Since \model\ utilizes the explicit point cloud representation to control the object deformation, our method relies on high quality point cloud representation to achieve fine-grained object deformations. 
The outlier points that are far away from the object surface can thus affect the quality of point cloud deformation, as it is hard to know which particular geometry region these outlier points are aligned with (e.g., outlier points above Lego bulldozer's blade, as shown in Figure \ref{fig:pcd_compare}). 
To improve the quality of point cloud representation, we purpose new point pruning and growing mechanisms for \model.

\subsection{Point pruning} 
The original Point-NeRF uses a learnable point confidence attribute to prune the low-confidence points, however, this pruning method does not guarantee the outlier points can be completely removed (see Figure \ref{fig:pcd_compare}).
In \model, we completely discard the point confidence attribute, and use estimated SDF values and recorded compositing weights to prune points. There are two key techniques in our pruning mechanism:

\textbf{Pruning based on estimated SDF.} We can easily compute per-point SDF values by treating a neural point's position as a query point's position (i.e., $\mathbf{x} = \mathbf{p}_i$). We then prune points with absolute SDF values larger than the pre-defined threshold (we use the SDF value $s$ such that $\frac{1}{\beta}\Psi_{\beta}(-s) = 0.99$ as the threshold).

\textbf{Pruning based on recorded compositing weights.}
The second pruning technique is based on an observation of compositing weights (Eqn. \ref{eq_radiance}), that is, 
if a query point $\mathbf{x}_t$  always has a very low compositing weight $w_t$ given a camera ray from any direction, then the region where this point locates is likely to be an empty space. 
We utilize this fact to prune neural points that are always associated with query points with low compositing weights.
Specifically, we keep a record of each neural point's maximum compositing weight that their associated query points can achieve during a fixed number of training iterations. We then prune points with recorded weights lower than a pre-defined threshold (we set the threshold to 0.02). After the pruning, we clear the recorded weights and start a new recording for the next round of pruning.

We jointly use the above two pruning methods to remove outlier points and random points inside the object. We only run the pruning in the NeRF training stage (Sec. \ref{sec:train_s1}), and the pruning is performed once for every 10k training iterations. As demonstrated in Figure \ref{fig:pcd_compare}, our purposed pruning mechanism can effectively remove the outliers that the original Point-NeRF is unable to remove.




\subsection{Point growing}

In addition to the original Point-NeRF's point growing mechanism, we also purpose a new point growing mechanism in order to add missing points to the complicated scenes.
Our new point growing mechanism works on the voxelized point cloud. Our method attempts to add new points to the neighbors of voxels that are very close to the surface (the estimated SDF values close to 0) but contain relatively few neighboring neural points. Please see Algorithm \ref{algo:growing} for more detailed steps of our point growing.

Similar to the point pruning mechanism, point growing is only applied to the NeRF training stage. We run our purposed point growing once for every 40k training iterations.
Figure \ref{fig:growing_compare} demonstrates the effectiveness of our purposed point growing mechanism. Compared to the Point-NeRF, our rendering results successfully recovers the missing points on the surface that is less observable.


\IncMargin{1.2em}
\begin{algorithm}
\caption{Our Proposed Point Growing\label{algo:growing}}
\SetKwInOut{Input}{Input}
\SetKwInOut{Output}{output}
\SetKwInOut{Define}{Def.}
\SetKwComment{Comment}{/* }{ */}
\Input{neural point cloud $\mathcal{P}$, SDF threshold $T_{\text{SDF}}$, point number threshold $T_{\text{points}}$, $K$, $MLP$}
\Define{$\cdot$ \texttt{neighbor}($V$): return a set of 26 adjacent neighboring voxels of voxel $V$.\newline
$\cdot$ \texttt{num\_points}($\mathcal{V}$): return the number of points contained by voxel set $\mathcal{V}$.\newline
$\cdot$ \texttt{topk\_min\_voxel}($\mathcal{V}$, $K$): return top $K$ voxels with the fewest points in $\mathcal{V}$.
}
Voxelize point cloud $\mathcal{P}$ to voxel set $\mathcal{V} =\{V_i\}$;\\
\For{$V_i$ in $\mathcal{V}$}{
    $s_i \leftarrow$ the estimated SDF of the voxel center $c_i$;\\
    $\mathcal{V}_{\text{local}} \leftarrow \{V_i\} \cup \texttt{neighbor}(V_i)$;\\
    $n_i \leftarrow n_i + \texttt{num\_points}(\mathcal{V}_{\text{local}}$);
}
$\mathcal{V} \leftarrow \{V_i \mid n_i < T_{\text{points}} \text{ and } |s_i| < T_{\text{SDF}}\}$;\\

\For{ $V_i$ in $\mathcal{V}$}{
    $\mathcal{V}_{\text{local}} \leftarrow \{V_i\} \cup \texttt{neighbor}(V_i)$;\\
    $\mathcal{V}_{\text{sparse}} \leftarrow \texttt{topk\_min\_voxel}(\mathcal{V}_{\text{local}}, K)$;\\
    $\mathcal{P} \leftarrow \mathcal{P}\cup \{c_i \mid V_i \in \mathcal{V}_{\text{sparse}}\}$ \Comment*{Add voxel centers}
}
return $\mathcal{P}$;
\end{algorithm}
\DecMargin{1.2em}
\vspace{-5pt}

%% file: appendices/additional_eval.tex
\begin{table*}[t]
    \vspace{-10pt}
    \centering
    \footnotesize
    \setlength\tabcolsep{0pt}
    \begin{tabularx}{\linewidth}%
    {*{6}{>{\centering\arraybackslash}X}}
    \multicolumn{6}{c}{\includegraphics[width=\linewidth]{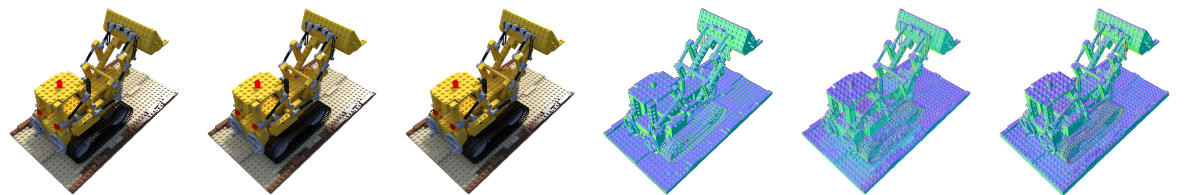}}%
    \\
    \multicolumn{6}{c}{\includegraphics[width=\linewidth]{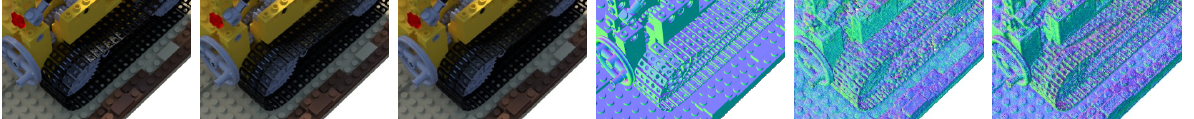}}%
    \\
    Ground truth & w/o $\mathcal{L}_d$ & w/ $\mathcal{L}_d$ & GT normal & Normal, w/o $\mathcal{L}_d$ & Normal, w/ $\mathcal{L}_d$
    \end{tabularx}%
    \makeatletter\def\@captype{figure}\makeatother
    \caption{Ablation on SDF continuity regularization $\mathcal{L}_d$ on Lego scene. We show the results of our non-BRDF model after 80k iterations.
    The estimated normals shown in the figure are SDF computed normals $\hat{\mathbf{n}}$. \label{tab:lego_ablation}}
    \vspace{-5pt}
\end{table*}

%% file: appendices/ablation.tex
\section{Ablation Studies}
\subsection{SDF regularizations}

Section \ref{sec:ablation} analyzes the effect of our proposed two SDF regularizations. We use another example in this subsection to show the necessity of our SDF regularization for learning accurate surface.


Figure \ref{tab:lego_ablation} shows how our proposed SDF continuity regularization $\mathcal{L}_d$ (Eqn. \ref{eq:l_d}) improves the quality of the estimated surface normals. 
Although we cannot observe significant differences in the synthesized images from models with and without the regularization $\mathcal{L}_d$, we do observe the differences in the estimated normals. The normal computed by the model without the regularization $\mathcal{L}_d$ is more likely to be affected by the texture colors/shadows from the observed training images and is unable to recover the geometry details in those shadowed regions (see the enlarged image patches in Fig. \ref{tab:lego_ablation}).
In contrast, the model with our SDF continuity regularization $\mathcal{L}_d$ is unaffected by the coloring or shadowing effects from training images and is able to show more consistent normal directions within a local region, which is essential to our lighting decomposition and BRDF-based rendering.

\subsection{Decomposition training loss}
\begin{figure*}[t]
    \vspace{-8pt}
    \centering
    \includegraphics[width=\linewidth]{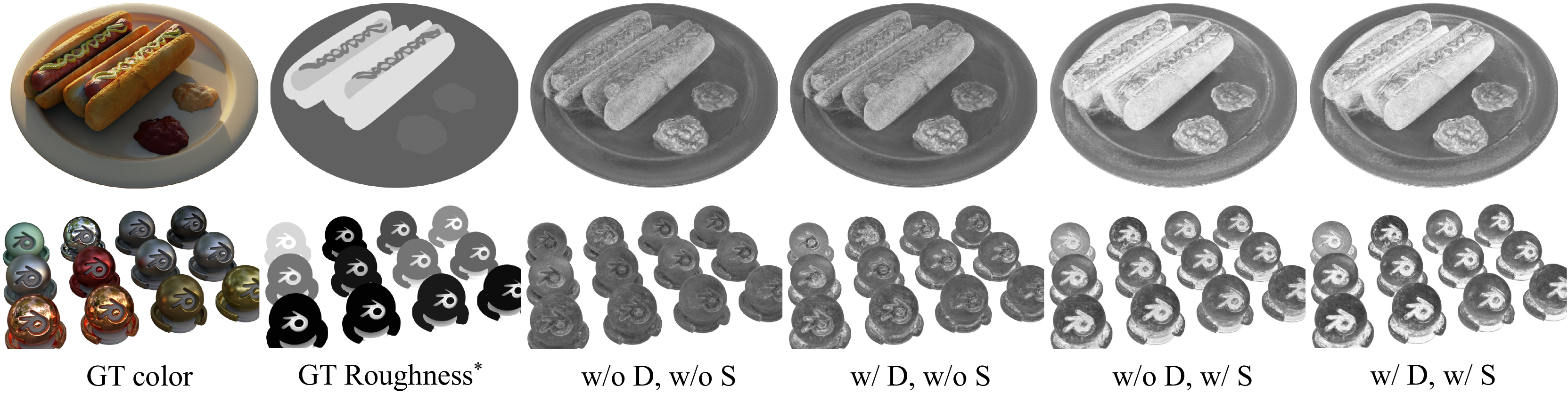}
    \vspace*{-20pt}
    \caption{
    The estimated material roughness under different light decomposition training strategies. `D' and `S' represent extra diffuse and extra specular color supervision, respectively. Note that the ground truth roughness is obtained by replacing the material base color with the corresponding roughness values in Blender's shader nodes. Since Blender has varied rendering settings for different materials, the ground truth roughness may have different meanings as opposed to our estimated roughness.
    }
    \label{fig:lighting_loss}
    \vspace{-5pt}
\end{figure*}

In Section \ref{sec:training_s2}, we shortly discuss the effect of using predicted specular color $\mathbf{C}_s$ to supervise our BRDF integrated specular color $\mathbf{C}_s^\prime$. 
The light training loss (Eqn. \ref{eq:loss_light}) shown in the main paper only additionally considers predicted specular color $\mathbf{C}_s$. In this subsection, we also take into account the predicted diffuse color $\mathbf{C}_d$ as one supervision term:
\begin{align}
    \mathcal{L}_{light} =& \lambda_{c} \mathcal{L}_{color}(\mathbf{C}^*, \mathbf{C}_d^\prime + \mathbf{C}_s^\prime) + \notag \\
    &\lambda_{g} \mathcal{L}_{color}(\mathbf{C}_s, \mathbf{C}_s^\prime) + \lambda_{l} \mathcal{L}_{color}(\mathbf{C}_d, \mathbf{C}_d^\prime)
    \label{eq:loss_light_ext}
\vspace{-9pt}
\end{align}
We justify our design choice by evaluating different combinations of color supervision during the light decomposition training.
Since learning BRDF parameters for specular reflectance is one of the major goals of this training, therefore, we use the estimated material roughness as the specular BRDF parameter for qualitative comparison.
We run the light decomposition training with 4 combinations of extra specular and extra diffuse color loss, and show the results in Figure \ref{fig:lighting_loss}.
When the models are trained without the extra specular color loss (labels containing ``w/o S"), the resulting roughness maps are unable to clearly distinguish the different materials.
Exclusively adding extra diffuse color loss (``w/ D, w/o S") can help separate the materials but the improvements are very limited (only observable from the Materials scene).
In contrast, the use of extra specular color loss (``w/ S") can effectively help \model\ to distinguish the different scene materials. The resulting roughness images are more like the ground truth roughness, which indicates a more accurate estimation of the material BRDF. 
Besides, jointly using both extra diffuse and specular color supervision does not provide significant improvements in the roughness estimation. Therefore, for simplicity, we only use the predicted specular color $\mathbf{C}_s$ for the extra training supervision (i.e., the original Eqn. \ref{eq:loss_light}).

%% file: appendices/additional_res.tex
\section{Additional Results}
\subsection{Per-scene breakdown}
Table \ref{tab:perscene_synthetic} shows the per-scene breakdown comparison of the quantitative results shown in Table \ref{tab:render_scores}. Our \model\ model has rendering scores comparable to the original Point-NeRF and has significantly better results on the estimated surface normals.
Table \ref{tab:deform_metrics} shows the comparison of rendering quality on all the evaluated deformed poses. \model\ performs consistently better than other methods on deformed scenes.

\subsection{Qualitative results of the reconstructed scenes}
In Figure \ref{tab:nerf_syn_result} and Figure \ref{tab:bmvs_result}, we present more visual results of the tested scenes from both synthetic and real-captured datasets.
In these figures, we visualize our estimated normal, roughness, visibility, and environment light. These are the key factors that control \model's illumination effects.
In addition to the scenes used for quantitative evaluations, we also test our model on two more synthetic scenes (Mannequin and Trex) and two challenging real-captured scenes from BlendedMVS (Evangelion and Gundam). Note that these two real scenes have unmasked backgrounds. In order to model the unbounded background, we employ a simplified NeRF++ \cites{zhang2020nerf++} model (fewer layers, and no view-dependent MLP) to separately represent the background. We use COLMAP \cites{schonberger2016pixelwise} to initialize the point cloud for these two real scenes. The qualitative results are shown in Figure \ref{tab:add_result}.
These qualitative results show that \model\ can make convincingly good estimations of surface normals as well as BRDF parameters for most scenes. The comparison between the ground truth environment light and our estimated environment light in synthetic scenes also demonstrates the effectiveness of our novel presentation of the environment light.
Another observation is that our estimated normal images are noisier in real-captured scenes compared to synthetic scenes. This is because the real-captured scenes have much more high-frequency details in their textures. To model these texture details, our discrete neural features have more variations in their parameter space, which further results in more variations in the estimated normal. Although VolSDF \cite{yariv2021volume} can get a smoother normal estimation, this is at the cost of lower rendering quality. Our predicted normal can be treated as a balance point between image rendering quality and normal/surface smoothness. %

\subsection{Additional editing results}
In addition to the editing results shown in the main paper, we showcase more editing results in this part. 
We choose other evaluated scenes to perform various geometry deformations and show the results in Figure \ref{tab:deform_result}. 
The results show that \model\ is able to achieve realistic rendering of deformed objects even in these challenging real-captured scenes.

We finally show some hybrid scene editings (deformation and relighting) to \model's represented scenes in Figure \ref{tab:relight}.
These rendering results show that \model\ is able to effectively synthesize illumination and shadows even for deformed objects. However, we should also note that \model\ cannot completely remove the shadows existing in the observed training images (e.g., \model\ does not remove the shadow in the Evangelion scene, we can see that unchanged shadow on the ground after relighting). Complete shadow removal from images under the same environment illumination is an extremely difficult task since shadows can be misinterpreted as the object's texture color. Future work on learning decomposed BRDF parameters from images under multiple environments (e.g., NeRD \cites{boss2021nerd} and Neural-PIL \cites{boss2021neural}) would help \model\ achieve better results on relighting.



\begin{table*}[hbt]
\caption{Per-scene quantitative results}
\input{tables/blender_synth_per_scene}
\vspace*{-5pt}
\label{tab:perscene_synthetic}
\end{table*}

\begin{table*}
    \caption{Quantitative evaluation of rendering the deformed scenes. The Avg. column shows the average scores over all the deformed poses excluding the original pose. \model$^*$ indicates our non-PBR rendering, while \model\ indicates our PBR rendering.
    }
    \label{tab:deform_metrics}
    \input{tables/deform_eval.tex}
\end{table*}

\begin{table*}[b]
    \centering
    \footnotesize
    \setlength\tabcolsep{0pt}
        \begin{tabularx}{\linewidth}%
        {p{1em}*{8}{>{\centering\arraybackslash}X}}
        \hdashline
        \rotatebox{90}{Ficus} & \multicolumn{8}{c}{\includegraphics[width=0.98\linewidth]{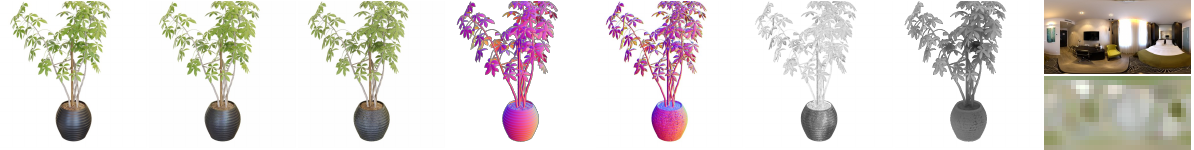}}\\ \hdashline
        \rotatebox{90}{Chair} & \multicolumn{8}{c}{\includegraphics[width=0.98\linewidth]{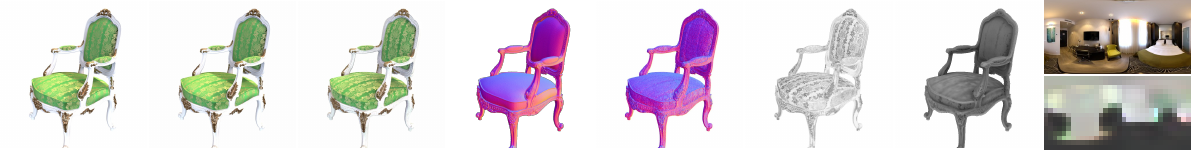}}\\ \hdashline
        \rotatebox{90}{Drums} & \multicolumn{8}{c}{\includegraphics[width=0.98\linewidth]{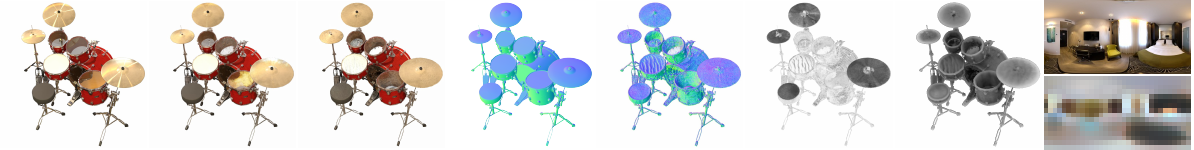}}\\ \hdashline
        \rotatebox{90}{Lego} & \multicolumn{8}{c}{\includegraphics[width=0.98\linewidth]{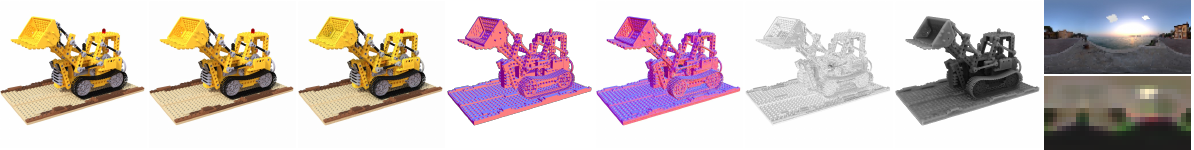}}\\ \hdashline
        \rotatebox{90}{Hotdog} & \multicolumn{8}{c}{\includegraphics[width=0.98\linewidth]{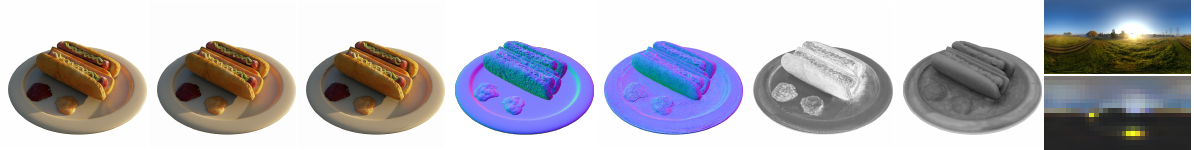}}\\ \hdashline
        \rotatebox{90}{Ship} & \multicolumn{8}{c}{\includegraphics[width=0.98\linewidth]{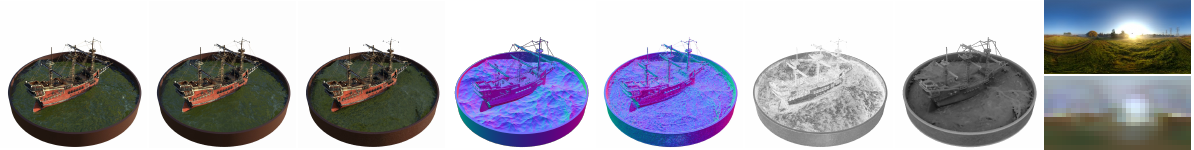}}\\ \hdashline
        \rotatebox{90}{Materials} & \multicolumn{8}{c}{\includegraphics[width=0.98\linewidth]{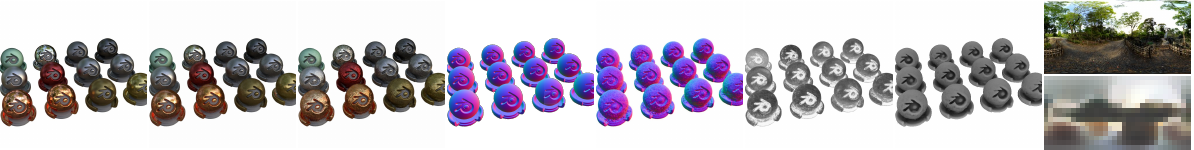}}\\ \hdashline
        \rotatebox{90}{Mic} & \multicolumn{8}{c}{\includegraphics[width=0.98\linewidth]{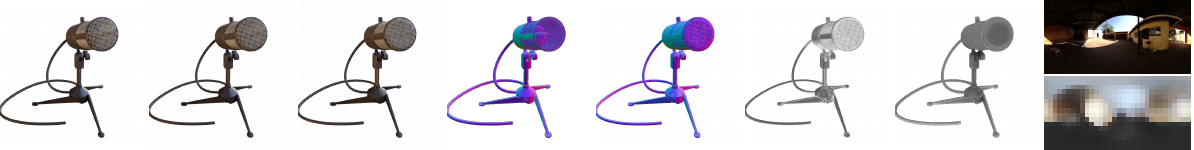}}\\ \hdashline
        \\
             & Ground truth & Our non-PBR & Our PBR & GT normal & Pred normal & Roughness & Visibility & Environment
        \end{tabularx}%
        
    \makeatletter\def\@captype{figure}\makeatother
    \caption{Qualitative results of the synthetic scenes. Note that in the ``Environment" column, the upper row shows the ground truth environment light, and the lower row shows our estimated environment light.
    \label{tab:nerf_syn_result}}
\end{table*}

\begin{table*}[b]
    \vspace{5pt}
    \centering
    \footnotesize
    \setlength\tabcolsep{0pt}
    \begin{tabularx}{\linewidth}%
    {p{1em}*{8}{>{\centering\arraybackslash}X}}
        \hdashline
        \rotatebox{90}{Character} & \multicolumn{8}{c}{\includegraphics[width=0.98\linewidth]{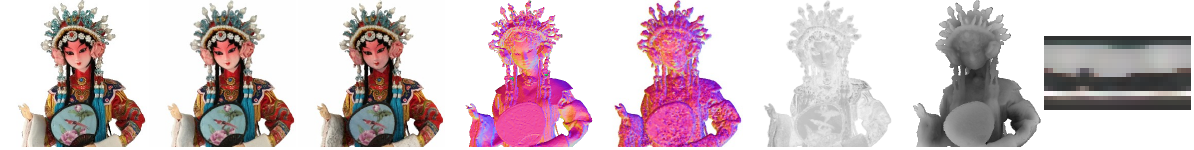}} \\ \hdashline
        \rotatebox{90}{Fountain} & \multicolumn{8}{c}{\includegraphics[width=0.98\linewidth]{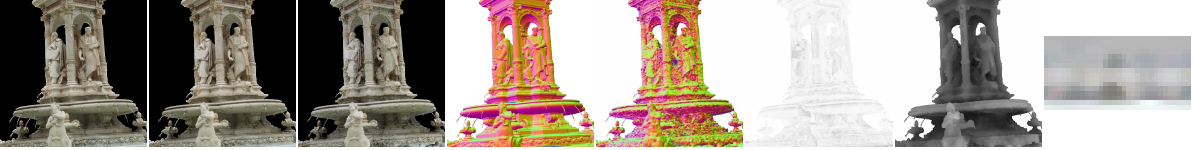}} \\ \hdashline
        \rotatebox{90}{Statues} & \multicolumn{8}{c}{\includegraphics[width=0.98\linewidth]{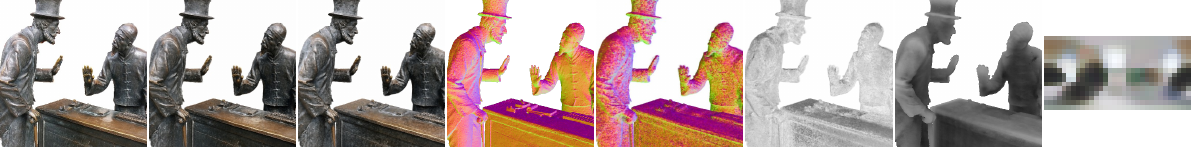}} \\ \hdashline
        \rotatebox{90}{Jade} & \multicolumn{8}{c}{\includegraphics[width=0.98\linewidth]{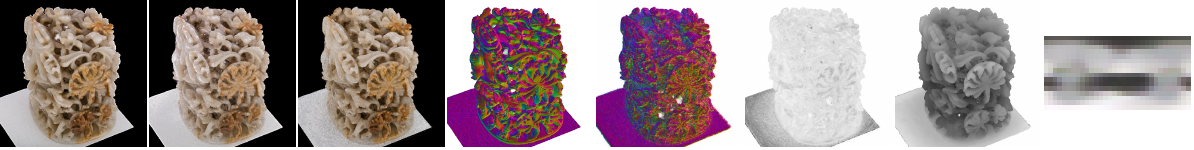}} \\ \hdashline
        \\
        & Ground truth & Our non-PBR & Our PBR & GT normal & Pred normal & Roughness & Visibility & Environment
    \end{tabularx}%
    \vspace*{-5pt}
    \makeatletter\def\@captype{figure}\makeatother
    \caption{Qualitative results of real-captured scenes from BlendedMVS dataset. Since these real-captured scenes do not have ground truth environment light, we only show our estimated environment light in the last column.\label{tab:bmvs_result}}
\end{table*}
\begin{table*}[b]
    \vspace{5pt}
    \centering
    \footnotesize
    \setlength\tabcolsep{0pt}
    \begin{tabularx}{\linewidth}%
    {p{1em}*{8}{>{\centering\arraybackslash}X}}
        \hdashline
        \rotatebox{90}{Trex} & \multicolumn{8}{c}{\includegraphics[width=0.98\linewidth]{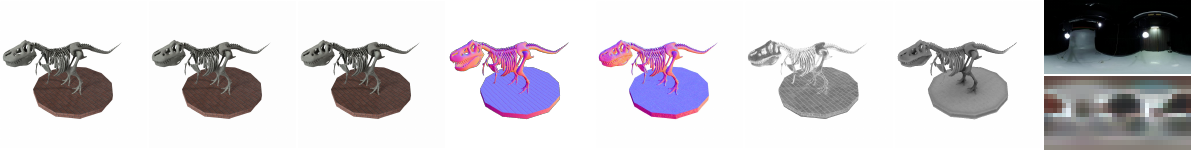}}\\ \hdashline
        \rotatebox{90}{Mannequin} & \multicolumn{8}{c}{\includegraphics[width=0.98\linewidth]{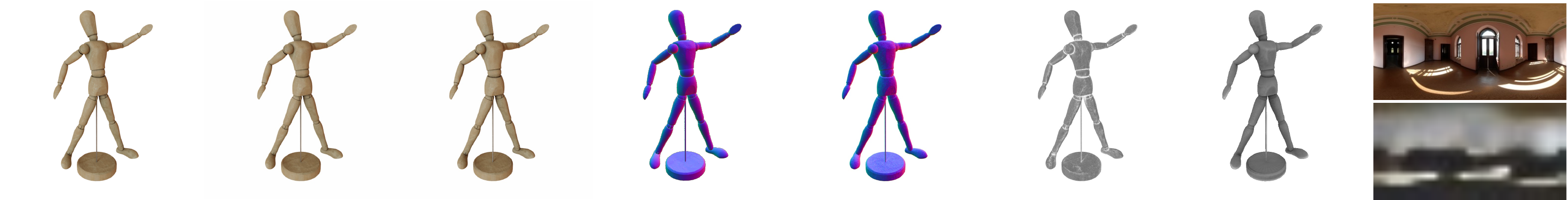}}\\ \hdashline
        \rotatebox{90}{Evangelion} & \multicolumn{8}{c}{\includegraphics[width=0.98\linewidth]{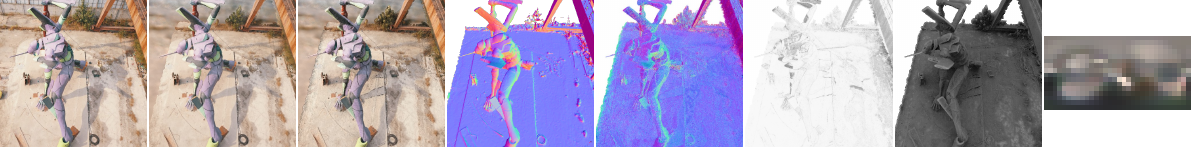}} \\ \hdashline
        \rotatebox{90}{Gundam} & \multicolumn{8}{c}{\includegraphics[width=0.98\linewidth]{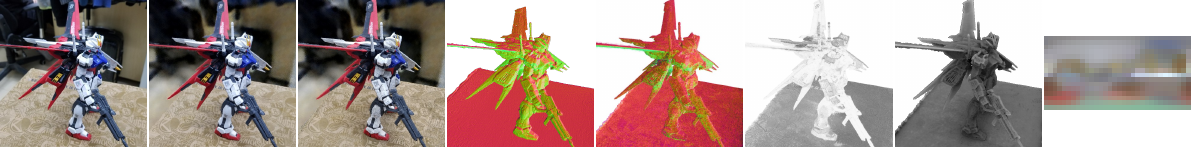}} \\ \hdashline
        \\
        & Ground truth & Our non-PBR & Our PBR & GT normal & Pred normal & Roughness & Visibility & Environment
    \end{tabularx}%
    \vspace*{-5pt}
    \makeatletter\def\@captype{figure}\makeatother
    \caption{Qualitative results of our additionally tested scenes. \label{tab:add_result}}
\end{table*}

\begin{table*}[t]
    \centering
        \footnotesize
        \setlength\tabcolsep{0pt}
        \begin{tabularx}{\linewidth}%
        {p{1em}*{4}{>{\centering\arraybackslash}X}}
            \rotatebox{90}{Character} & \multicolumn{4}{c}{\includegraphics[width=0.98\linewidth]{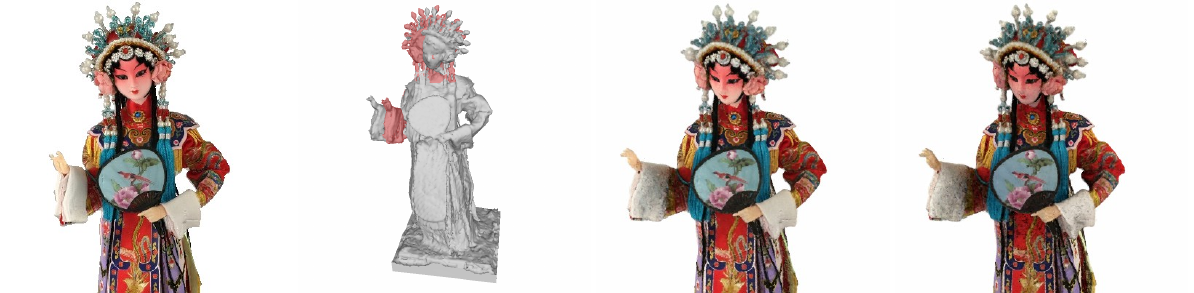}}\\
            \rotatebox{90}{Statues} & \multicolumn{4}{c}{\includegraphics[width=0.98\linewidth]{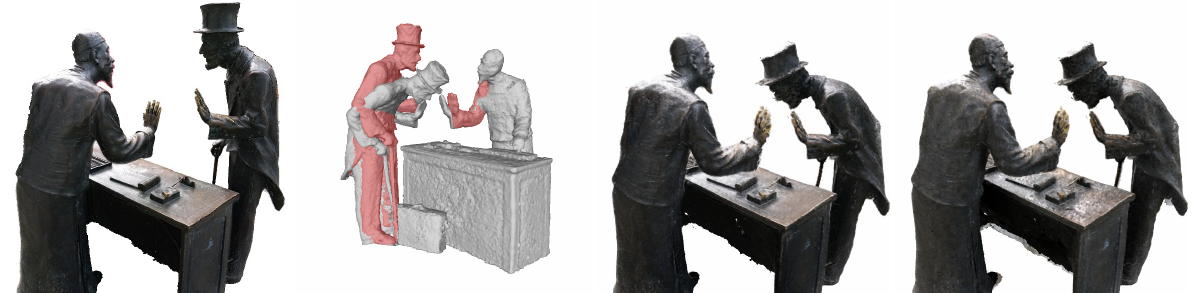}}\\
            \rotatebox{90}{Trex} & \multicolumn{4}{c}{\includegraphics[width=0.98\linewidth]{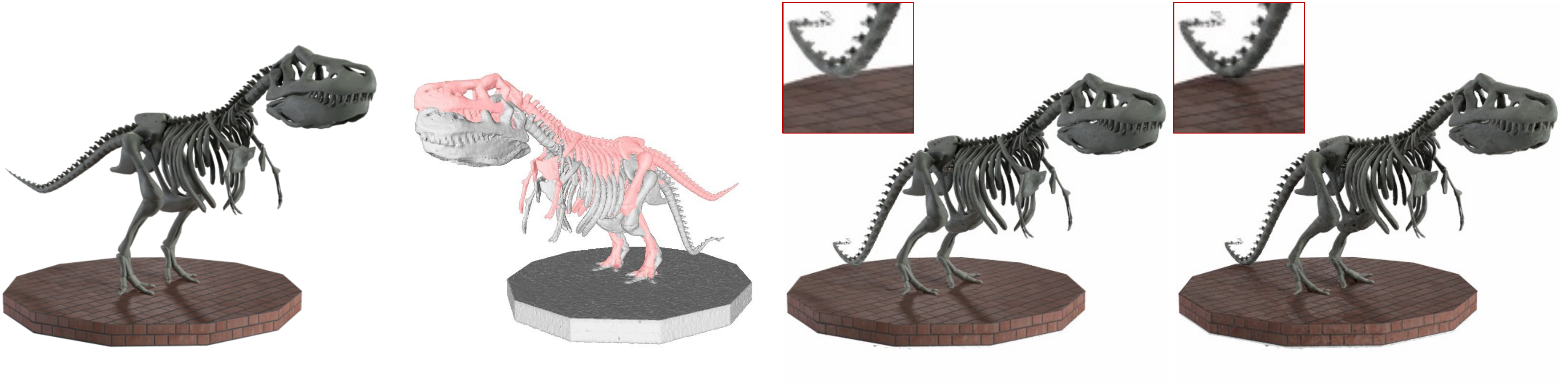}}\\
            \rotatebox{90}{Evangelion} & \multicolumn{4}{c}{\includegraphics[width=0.98\linewidth]{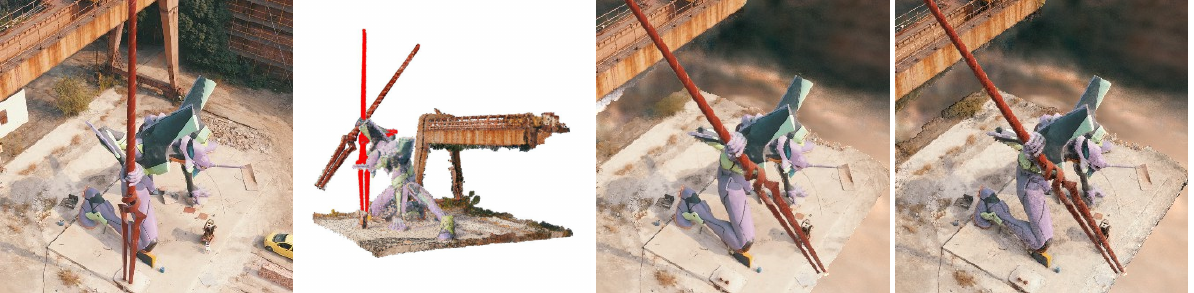}}\\
            \\
            \rotatebox{90}{Gundam} & \multicolumn{4}{c}{\includegraphics[width=0.98\linewidth]{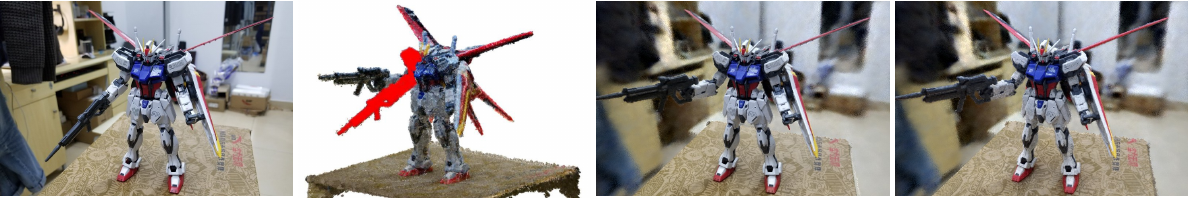}}\\
            & Ground truth & Editing & Our non-PBR & Our PBR 
        \end{tabularx}%
    \makeatletter\def\@captype{figure}\makeatother
    \caption{
    Qualitative results on geometry editing. We perform the ARAP mesh deformations in the first 3 scenes and direct point manipulation in the last 2 scenes. The meshes shown in the first 3 scenes are extracted by marching cube on the zero-level set of \model's estimated SDF.\label{tab:deform_result}}
\end{table*}

\begin{table*}[t]
    \centering
        \footnotesize
        \setlength\tabcolsep{0pt}
        \begin{tabularx}{\linewidth}%
        {p{1em}*{2}{>{\centering\arraybackslash}X}*{2}{>{\centering\arraybackslash}p{0.3\linewidth}}}
            \rotatebox{90}{\;Materials} & \multicolumn{4}{c}{\includegraphics[width=0.98\linewidth]{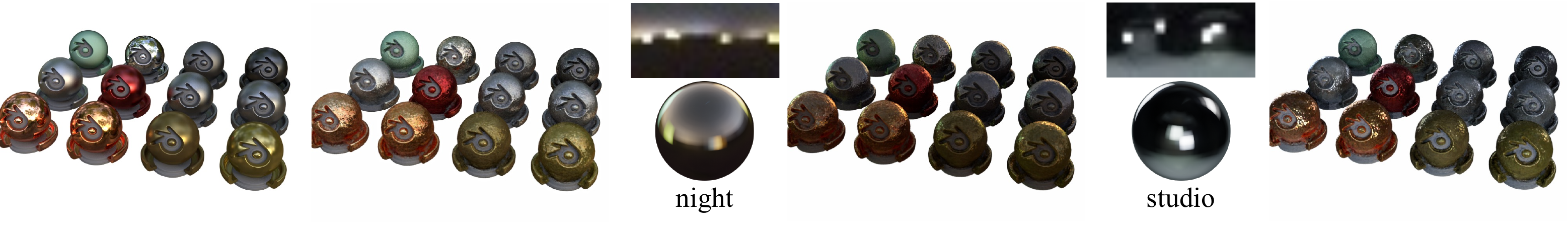}}\\
            \rotatebox{90}{\;Trex} & \multicolumn{4}{c}{\includegraphics[width=0.98\linewidth]{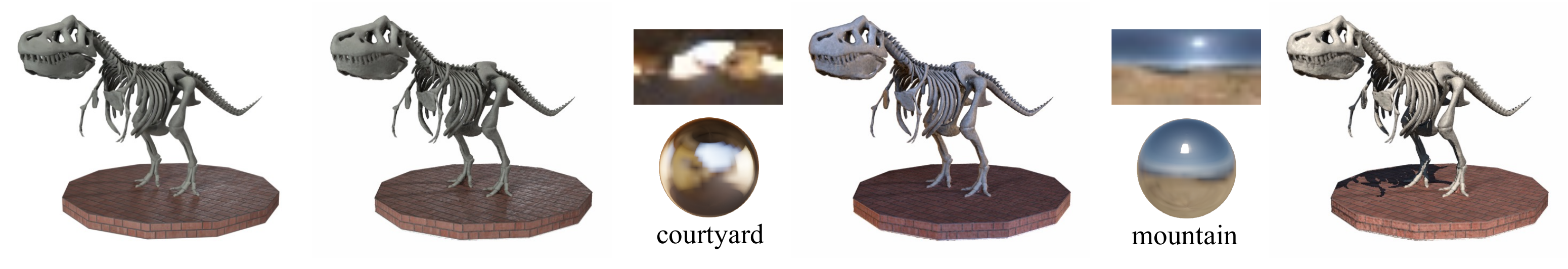}}\\
            \rotatebox{90}{\;Evangelion} & \multicolumn{4}{c}{\includegraphics[width=0.98\linewidth]{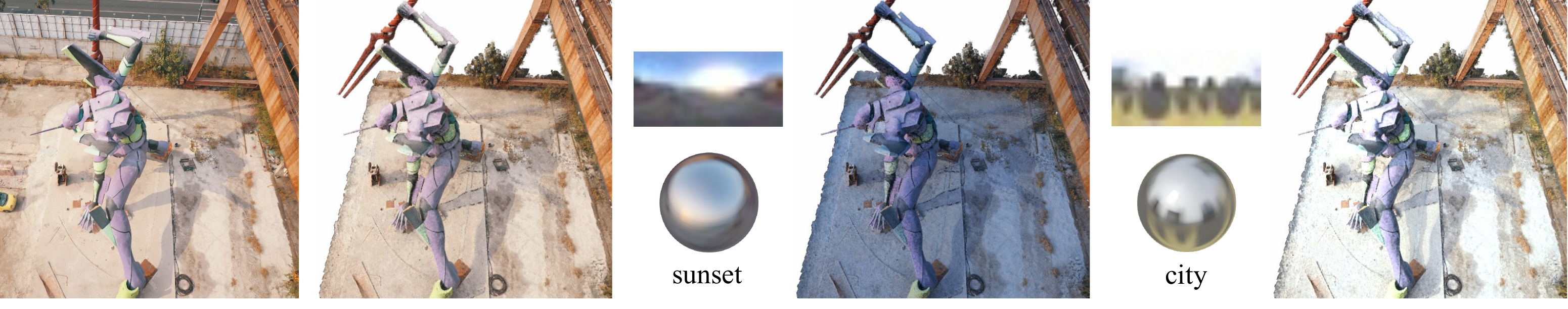}}\\
            \rotatebox{90}{\;Gundam} & \multicolumn{4}{c}{\includegraphics[width=0.98\linewidth]{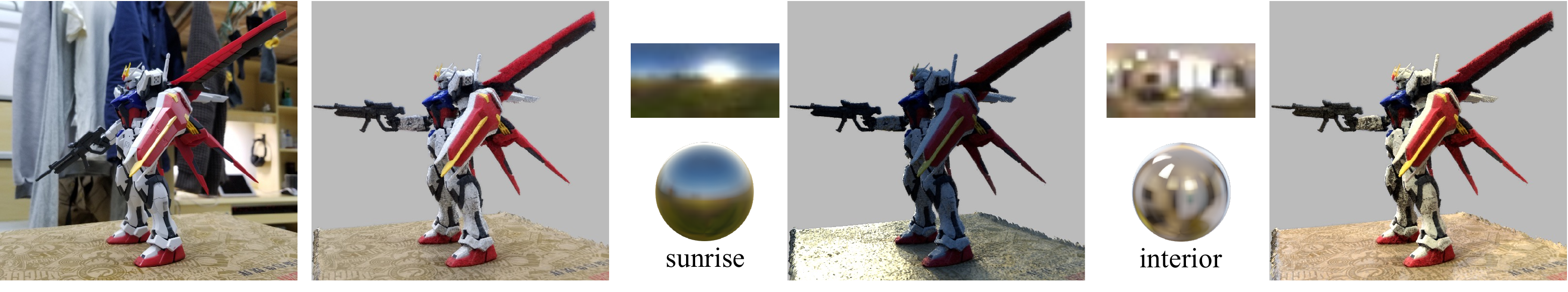}}\\
            & Ground truth & Our PBR synthesis & Relighting \#1 & Relighting \#2
        \end{tabularx}%
    \makeatletter\def\@captype{figure}\makeatother
    \caption{
    Scene relighting results.\label{tab:relight}}
\end{table*}

%% file: tables/blender_synth_per_scene.tex
\small
\setlength\tabcolsep{0pt}
\begin{tabularx}{\linewidth}
    {{>{\arraybackslash}p{1em}|}*{8}{>{\centering\arraybackslash}X}|*{4}{>{\centering\arraybackslash}X}}
\hline
 &\multicolumn{8}{c|}{NeRF Synthetic} &\multicolumn{4}{c}{Blended MVS} \\ 
 & Chair & Drums & Lego  & Mic & Materials & Ship & Hotdog & Ficus & Jade & Fountain & Character & Statues \\ \hline
\multicolumn{13}{c}{\textbf{PSNR $\uparrow$}}                                                                   \\ \hline
\multicolumn{1}{l|}{NSVF}      & 33.00 & 25.18 & 32.54 & 34.27 & \cellcolor{orange!25}32.68     & 27.93 & \cellcolor{yellow!25}37.14  & 31.23 & \cellcolor{orange!25}26.96 & \cellcolor{orange!25}27.73    & \cellcolor{yellow!25}27.95     & 24.97   \\
\multicolumn{1}{l|}{VolSDF}    & 30.57 & 20.43 & 29.46 & 30.53 & \cellcolor{yellow!25}29.13     & 25.51 & 35.11  & 22.91 & 25.20 & 24.05    & 25.59     & 24.26   \\
\multicolumn{1}{l|}{NVDiffRec} & 31.60 & 24.10 & 29.14 & 30.78 & 26.74     & 26.12 & 33.04  & 30.88 &  &     &      &    \\
\multicolumn{1}{l|}{PointNeRF} & \cellcolor{orange!25}35.40 & \cellcolor{orange!25}26.06 & \cellcolor{orange!25}35.04 & \cellcolor{orange!25}35.95 & 29.61     & \cellcolor{orange!25}30.97 & \cellcolor{orange!25}37.30  & \cellcolor{orange!25}36.13 & \cellcolor{yellow!25}26.14 & 25.68    & \cellcolor{orange!25}29.06     & \cellcolor{orange!25}25.97   \\
\multicolumn{1}{l|}{SPIDR$^*$}      & \cellcolor{yellow!25}34.84 & \cellcolor{yellow!25}25.70 & \cellcolor{yellow!25}34.77 & \cellcolor{yellow!25}34.47 & 27.42     & \cellcolor{yellow!25}30.74 & 36.82  & \cellcolor{yellow!25}33.66 & 25.97 & \cellcolor{yellow!25}27.25    & 27.80     & \cellcolor{yellow!25}25.50   \\
\multicolumn{1}{l|}{SPIDR} & 30.33 & 23.21 & 29.67 & 27.88 & 23.74     & 26.52 & 32.72  & 28.21 & 25.00 & 24.44    & 25.42     & 23.95   \\
\hline
\multicolumn{13}{c}{\textbf{SSIM $\uparrow$}}                                                                   \\ \hline
\multicolumn{1}{l|}{NSVF}      & 0.968 & 0.931 & 0.960 & 0.987 & \cellcolor{orange!25}0.973     & 0.854 & \cellcolor{yellow!25}0.980  & 0.973 & 0.901 & 0.913    & 0.921     & 0.858   \\
\multicolumn{1}{l|}{VolSDF}    & 0.949 & 0.893 & 0.951 & 0.969 & 0.954     & 0.842 & 0.972  & 0.929 & 0.919 & 0.908    & 0.939     & 0.895   \\
\multicolumn{1}{l|}{NVDiffRec} & 0.969 & 0.916 & 0.949 & 0.977 & 0.923     & 0.833 & 0.973  & 0.970 &  &     &      &    \\
\multicolumn{1}{l|}{PointNeRF} & \cellcolor{orange!25}0.991 & \cellcolor{orange!25}0.954 & \cellcolor{yellow!25}0.988 & \cellcolor{orange!25}0.994 & \cellcolor{yellow!25}0.971     & \cellcolor{yellow!25}0.942 & \cellcolor{orange!25}0.991  & \cellcolor{orange!25}0.993 & \cellcolor{yellow!25}0.931 & \cellcolor{yellow!25}0.935    & \cellcolor{orange!25}0.970     & \cellcolor{yellow!25}0.930   \\
\multicolumn{1}{l|}{SPIDR$^*$}      & \cellcolor{orange!25}0.991 & \cellcolor{yellow!25}0.957 & \cellcolor{orange!25}0.989 & \cellcolor{yellow!25}0.993 & 0.952     & \cellcolor{orange!25} 0.945 & \cellcolor{orange!25}0.991  & \cellcolor{yellow!25}0.990 & \cellcolor{orange!25}0.935 & \cellcolor{orange!25}0.957    & \cellcolor{yellow!25}0.968     & \cellcolor{orange!25}0.933   \\
\multicolumn{1}{l|}{SPIDR} & \cellcolor{yellow!25}0.974 & 0.929 & 0.969 & 0.976 & 0.909     & 0.897 & 0.979  & 0.970 & 0.898 & 0.922    & 0.950     & 0.910   \\
\hline
\multicolumn{13}{c}{\textbf{LPIPS $\downarrow$}}                                                                  \\ \hline
\multicolumn{1}{l|}{NSVF}      & 0.043 & 0.069 & 0.029 & 0.010 & \cellcolor{orange!25}0.021     & 0.162 & 0.025  & 0.017 & 0.094 & 0.113    & 0.074     & 0.171   \\
\multicolumn{1}{l|}{VolSDF}    & 0.056 & 0.119 & 0.054 & 0.191 & 0.048     & 0.191 & 0.043  & 0.068 & 0.128 & 0.177    & 0.091     & 0.196   \\
\multicolumn{1}{l|}{NVDiffRec} & 0.045 & 0.101 & 0.061 & 0.040 & 0.100     & 0.191 & 0.060  & 0.048 &  &     &      &    \\
\multicolumn{1}{l|}{PointNeRF} & \cellcolor{orange!25}0.010 & \cellcolor{yellow!25}0.055 & \cellcolor{yellow!25}0.011 & \cellcolor{orange!25}0.007 & \cellcolor{yellow!25}0.041     & \cellcolor{orange!25}0.070 & \cellcolor{yellow!25}0.016  & \cellcolor{orange!25}0.009 & \cellcolor{yellow!25}0.091 & 0.104    & \cellcolor{orange!25}0.033     & \cellcolor{orange!25}0.082   \\
\multicolumn{1}{l|}{SPIDR$^*$}      & \cellcolor{orange!25}0.010 & \cellcolor{orange!25}0.049 & \cellcolor{orange!25}0.010 & \cellcolor{yellow!25}0.008 & 0.046     & \cellcolor{yellow!25}0.079 & \cellcolor{orange!25}0.014  & \cellcolor{yellow!25}0.015 & \cellcolor{orange!25}0.079 & \cellcolor{orange!25}0.061    & \cellcolor{yellow!25}0.039     & \cellcolor{yellow!25}0.087   \\
\multicolumn{1}{l|}{SPIDR} & \cellcolor{yellow!25}0.037 & 0.085 & 0.029 & 0.037 & 0.089     & 0.127 & 0.043  & 0.040 & 0.126 & \cellcolor{yellow!25}0.098    & 0.055     & 0.115   \\
\hline
\multicolumn{13}{c}{\textbf{MAE\si{\degree} $\downarrow$}}                                                               \\ \hline
\multicolumn{1}{l|}{VolSDF}    & \cellcolor{orange!25}14.09 & \cellcolor{orange!25}21.46 & 26.62 & \cellcolor{orange!25}19.58 & \cellcolor{orange!25}8.28  & \cellcolor{orange!25}16.97 & \cellcolor{orange!25}12.17 & \cellcolor{yellow!25}39.80 & \cellcolor{orange!25}32.17 & \cellcolor{orange!25}24.87 & \cellcolor{orange!25}28.74 & \cellcolor{orange!25} 18.03 \\
\multicolumn{1}{l|}{NVDiffRec}   & 
\cellcolor{yellow!25}20.26 & \cellcolor{yellow!25}30.72 & 36.59 & \cellcolor{yellow!25}20.84 & \cellcolor{yellow!25}10.78 & 33.53 & \cellcolor{yellow!25}17.63 & \cellcolor{orange!25}29.69 
\\
\multicolumn{1}{l|}{PointNeRF} & 37.24 & 54.13 & 40.97 & 47.51 & 60.41 & 50.82 & 32.61 & 61.01 & 50.90 & 44.59 & 58.30 & 54.61 \\
\multicolumn{1}{l|}{Ours-Grad} & 26.58 & 47.93 & \cellcolor{yellow!25}25.46 & 26.15 & 27.90 & 27.72 & 20.56 & 39.92 & 50.52 & 50.07 & 50.43 & 38.89 \\
\multicolumn{1}{l|}{Ours-Pred} & 22.19 & 39.86 & \cellcolor{orange!25}24.05 & 22.06 & 22.26 & \cellcolor{yellow!25}23.03 & 18.23 & 41.95 & \cellcolor{yellow!25}43.32 & \cellcolor{yellow!25}34.42 & \cellcolor{yellow!25}33.96 & \cellcolor{yellow!25}27.42 \\
\hline
\end{tabularx}

%% file: tables/deform_eval.tex
\small
\centering
\setlength\tabcolsep{5pt}
\begin{tabular}{l|c|ccccccc|c|c|cc|c}
\hline
\multicolumn{1}{l|}{}          & \multicolumn{9}{c|}{\textbf{Mannequin}}                                                                                                     & \multicolumn{4}{c}{\textbf{T-Rex}}                                         \\ \cline{2-14} 
\multicolumn{1}{c|}{}          & \multicolumn{1}{c|}{\textbf{Ori.}}  & Pose 1 & Pose 2 & Pose 3 & Pose 4 & Pose 5 & Pose 6 & \multicolumn{1}{c|}{Pose 7} & \multicolumn{1}{c|}{\textbf{Avg.}}   & \multicolumn{1}{c|}{\textbf{Ori.}}  & Pose 1 & \multicolumn{1}{c|}{Pose 2} & \textbf{Avg.}   \\ \hline
\multicolumn{1}{c}{}           & \multicolumn{13}{c}{\textbf{PSNR$\uparrow$}}                                                                                                                                                                                    \\ \hline
\multicolumn{1}{l|}{Xu \etal \cite{xu2022deforming}} & \cellcolor{orange!25}44.57 & 24.35 &	23.77 &	24.44 &	24.37 &	24.09 &	24.75 &	24.51  & 24.32 & \cellcolor{orange!25}40.06 &	25.02 &	23.10 &	24.06\\ 
\multicolumn{1}{l|}{NVDiffRec} &  37.28 &	28.62 &	26.45 &	28.13 &	29.23 &	27.80 &	28.67 &	29.18 & 28.30 & 31.49 &	28.47 &	24.86 &	26.67\\
\multicolumn{1}{l|}{SPIDR$^*$} & \cellcolor{yellow!25}40.68 & \cellcolor{yellow!25}31.04 &	\cellcolor{yellow!25}30.77 &	\cellcolor{yellow!25}30.23 &	\cellcolor{yellow!25}32.40 &	\cellcolor{yellow!25}33.15 &	\cellcolor{yellow!25}31.67 &	\cellcolor{yellow!25}30.49 &	\cellcolor{yellow!25}31.39 & \cellcolor{yellow!25}38.30 &	\cellcolor{orange!25}35.06 &	\cellcolor{yellow!25}31.77 &	\cellcolor{yellow!25}33.42\\
\multicolumn{1}{l|}{SPIDR}     & 39.78 & \cellcolor{orange!25}32.97 &	\cellcolor{orange!25}31.49 &	\cellcolor{orange!25}31.97 &	\cellcolor{orange!25}34.41 &	\cellcolor{orange!25}33.48 &	\cellcolor{orange!25}33.37 &	\cellcolor{orange!25}32.79 &	\cellcolor{orange!25}32.93 & 36.32 &	\cellcolor{yellow!25}34.38 &	\cellcolor{orange!25}32.55 &	\cellcolor{orange!25}33.47 \\ \hline
\multicolumn{1}{c}{}           & \multicolumn{13}{c}{\textbf{SSIM$\uparrow$}}                                                                                                                                                                                     \\ \hline
\multicolumn{1}{l|}{Xu \etal \cite{xu2022deforming}} & \cellcolor{orange!25}0.999 &	0.952 &	0.947 &	0.950 &	0.953 &	0.948 &	0.954 &	0.952 &	0.951 & \cellcolor{orange!25}0.996 &	0.944 &	0.939 &	0.941 \\ 
\multicolumn{1}{l|}{NVDiffRec} & 0.995 &	0.974 &	0.964 &	0.972 &	0.976 &	0.970 &	0.974 &	0.976 &	0.972 & 0.986 &	0.956 &	\cellcolor{yellow!25}0.942 &	0.949 \\
\multicolumn{1}{l|}{SPIDR$^*$} & \cellcolor{yellow!25}0.998 &	\cellcolor{yellow!25}0.986 &	\cellcolor{yellow!25}0.986 &	\cellcolor{yellow!25}0.985 &	\cellcolor{yellow!25}0.989 &	\cellcolor{yellow!25}0.991 &	\cellcolor{yellow!25}0.988 &	\cellcolor{yellow!25}0.985 &	\cellcolor{yellow!25}0.987 & \cellcolor{yellow!25}0.997 &	\cellcolor{orange!25}0.992 &	\cellcolor{orange!25}0.986 &	\cellcolor{orange!25}0.989 \\
\multicolumn{1}{l|}{SPIDR}     & 0.997 &	\cellcolor{orange!25}0.990 &	\cellcolor{orange!25}0.988 &	\cellcolor{orange!25}0.989 &	\cellcolor{orange!25}0.992 &	\cellcolor{orange!25}0.992 &	\cellcolor{orange!25}0.991 &	\cellcolor{orange!25}0.990 &	\cellcolor{orange!25}0.990 & 0.993 &	\cellcolor{yellow!25}0.989 &	\cellcolor{orange!25}0.986 &	\cellcolor{yellow!25}0.987 \\ \hline
\multicolumn{1}{c}{}           & \multicolumn{13}{c}{\textbf{LPIPS$\downarrow$}}                                                                                                                                                                                  \\ \hline
\multicolumn{1}{l|}{Xu \etal \cite{xu2022deforming}} & \cellcolor{orange!25}0.003 &	0.064 &	0.060 &	0.058 &	0.061 &	0.056 &	0.057 &	0.066 &	0.060 & \cellcolor{yellow!25}0.005 &	0.032 &	0.041 &	0.037\\ 
\multicolumn{1}{l|}{NVDiffRec} & 0.012 &	0.022 &	0.032 &	0.025 &	0.021 &	\cellcolor{yellow!25}0.025 &	0.023 &	0.022 &	0.024 & 0.025 &	0.040 &	0.047 &	0.044\\
\multicolumn{1}{l|}{SPIDR$^*$} & \cellcolor{orange!25}0.003 &	\cellcolor{yellow!25}0.017 &	\cellcolor{yellow!25}0.020 &	\cellcolor{yellow!25}0.018 &	\cellcolor{yellow!25}0.014 &	\cellcolor{orange!25}0.014 &	\cellcolor{yellow!25}0.016 &	\cellcolor{yellow!25}0.017 &	\cellcolor{yellow!25}0.016 & \cellcolor{orange!25}0.004 &	\cellcolor{orange!25}0.008 &	\cellcolor{orange!25}0.011 &	\cellcolor{orange!25}0.010\\
\multicolumn{1}{l|}{SPIDR}     & \cellcolor{yellow!25}0.005 &	\cellcolor{orange!25}0.015 &	\cellcolor{orange!25}0.019 &	\cellcolor{orange!25}0.016 &	\cellcolor{orange!25}0.012 &	\cellcolor{orange!25}0.014 &	\cellcolor{orange!25}0.014 &	\cellcolor{orange!25}0.016 &	\cellcolor{orange!25}0.015 & 0.008 &	\cellcolor{yellow!25}0.012 &	\cellcolor{yellow!25}0.014 &	\cellcolor{yellow!25}0.013 \\ \hline
\end{tabular}

%% file: PaperForReview.bbl
\begin{thebibliography}{10}\itemsep=-1pt

\bibitem{aittala2016reflectance}
Miika Aittala, Timo Aila, and Jaakko Lehtinen.
\newblock Reflectance modeling by neural texture synthesis.
\newblock {\em ACM Transactions on Graphics (ToG)}, 35(4):1--13, 2016.

\bibitem{anderson1996proposal}
Matthew Anderson, Ricardo Motta, Srinivasan Chandrasekar, and Michael Stokes.
\newblock Proposal for a standard default color space for the internet—srgb.
\newblock In {\em Color and imaging conference}, volume 1996, pages 238--245.
  Society for Imaging Science and Technology, 1996.

\bibitem{neumesh}
Chong Bao, Bangbang Yang, Zeng Junyi, Bao Hujun, Zhang Yinda, Cui Zhaopeng, and
  Zhang Guofeng.
\newblock Neumesh: Learning disentangled neural mesh-based implicit field for
  geometry and texture editing.
\newblock In {\em European Conference on Computer Vision (ECCV)}, 2022.

\bibitem{barron2021mip}
Jonathan~T Barron, Ben Mildenhall, Matthew Tancik, Peter Hedman, Ricardo
  Martin-Brualla, and Pratul~P Srinivasan.
\newblock Mip-nerf: A multiscale representation for anti-aliasing neural
  radiance fields.
\newblock In {\em Proceedings of the IEEE/CVF International Conference on
  Computer Vision}, pages 5855--5864, 2021.

\bibitem{barron2022mip}
Jonathan~T Barron, Ben Mildenhall, Dor Verbin, Pratul~P Srinivasan, and Peter
  Hedman.
\newblock Mip-nerf 360: Unbounded anti-aliased neural radiance fields.
\newblock In {\em Proceedings of the IEEE/CVF Conference on Computer Vision and
  Pattern Recognition}, pages 5470--5479, 2022.

\bibitem{bi2020neural}
Sai Bi, Zexiang Xu, Pratul Srinivasan, Ben Mildenhall, Kalyan Sunkavalli,
  Milo{\v{s}} Ha{\v{s}}an, Yannick Hold-Geoffroy, David Kriegman, and Ravi
  Ramamoorthi.
\newblock Neural reflectance fields for appearance acquisition.
\newblock {\em arXiv preprint arXiv:2008.03824}, 2020.

\bibitem{boss2021nerd}
Mark Boss, Raphael Braun, Varun Jampani, Jonathan~T Barron, Ce Liu, and Hendrik
  Lensch.
\newblock Nerd: Neural reflectance decomposition from image collections.
\newblock In {\em Proceedings of the IEEE/CVF International Conference on
  Computer Vision}, pages 12684--12694, 2021.

\bibitem{boss2021neural}
Mark Boss, Varun Jampani, Raphael Braun, Ce Liu, Jonathan Barron, and Hendrik
  Lensch.
\newblock Neural-pil: Neural pre-integrated lighting for reflectance
  decomposition.
\newblock {\em Advances in Neural Information Processing Systems},
  34:10691--10704, 2021.

\bibitem{chen2021mvsnerf}
Anpei Chen, Zexiang Xu, Fuqiang Zhao, Xiaoshuai Zhang, Fanbo Xiang, Jingyi Yu,
  and Hao Su.
\newblock Mvsnerf: Fast generalizable radiance field reconstruction from
  multi-view stereo.
\newblock In {\em Proceedings of the IEEE/CVF International Conference on
  Computer Vision}, pages 14124--14133, 2021.

\bibitem{debevec1998rendering}
Paul Debevec.
\newblock Rendering synthetic objects into real scenes: Bridging traditional
  and image-based graphics with global illumination and high dynamic range
  photography.
\newblock In {\em Proc. SIGGRAPH 98}, pages 189--198, 1998.

\bibitem{dellaert2020neural}
Frank Dellaert and Lin Yen-Chen.
\newblock Neural volume rendering: Nerf and beyond.
\newblock {\em arXiv preprint arXiv:2101.05204}, 2020.

\bibitem{deschaintre2018single}
Valentin Deschaintre, Miika Aittala, Fredo Durand, George Drettakis, and Adrien
  Bousseau.
\newblock Single-image svbrdf capture with a rendering-aware deep network.
\newblock {\em ACM Transactions on Graphics (ToG)}, 37(4):1--15, 2018.

\bibitem{eisemann2013efficient}
Elmar Eisemann, Ulf Assarsson, Michael Schwarz, Michal Valient, and Michael
  Wimmer.
\newblock Efficient real-time shadows.
\newblock In {\em ACM SIGGRAPH 2013 Courses}, pages 1--54. 2013.

\bibitem{ericson2004real}
Christer Ericson.
\newblock {\em Real-time collision detection}.
\newblock Crc Press, 2004.

\bibitem{fridovich2022plenoxels}
Sara Fridovich-Keil, Alex Yu, Matthew Tancik, Qinhong Chen, Benjamin Recht, and
  Angjoo Kanazawa.
\newblock Plenoxels: Radiance fields without neural networks.
\newblock In {\em Proceedings of the IEEE/CVF Conference on Computer Vision and
  Pattern Recognition}, pages 5501--5510, 2022.

\bibitem{guo2020object}
Michelle Guo, Alireza Fathi, Jiajun Wu, and Thomas Funkhouser.
\newblock Object-centric neural scene rendering.
\newblock {\em arXiv preprint arXiv:2012.08503}, 2020.

\bibitem{hedman2021baking}
Peter Hedman, Pratul~P Srinivasan, Ben Mildenhall, Jonathan~T Barron, and Paul
  Debevec.
\newblock Baking neural radiance fields for real-time view synthesis.
\newblock In {\em Proceedings of the IEEE/CVF International Conference on
  Computer Vision}, pages 5875--5884, 2021.

\bibitem{jacobson2012fast}
Alec Jacobson, Ilya Baran, Ladislav Kavan, Jovan Popovi{\'c}, and Olga Sorkine.
\newblock Fast automatic skinning transformations.
\newblock {\em ACM Transactions on Graphics (TOG)}, 31(4):1--10, 2012.

\bibitem{ju2005mean}
Tao Ju, Scott Schaefer, and Joe Warren.
\newblock Mean value coordinates for closed triangular meshes.
\newblock In {\em ACM Siggraph 2005 Papers}, pages 561--566. 2005.

\bibitem{kajiya1986rendering}
James~T Kajiya.
\newblock The rendering equation.
\newblock In {\em Proceedings of the 13th annual conference on Computer
  graphics and interactive techniques}, pages 143--150, 1986.

\bibitem{screened}
Michael Kazhdan and Hugues Hoppe.
\newblock Screened poisson surface reconstruction.
\newblock {\em ACM Transactions on Graphics (ToG)}, 32(3):1--13, 2013.

\bibitem{kholgade20143d}
Natasha Kholgade, Tomas Simon, Alexei Efros, and Yaser Sheikh.
\newblock 3d object manipulation in a single photograph using stock 3d models.
\newblock {\em ACM Transactions on Graphics (TOG)}, 33(4):1--12, 2014.

\bibitem{kingma2014adam}
Diederik~P Kingma and Jimmy Ba.
\newblock Adam: A method for stochastic optimization.
\newblock {\em arXiv preprint arXiv:1412.6980}, 2014.

\bibitem{lazova2022control}
Verica Lazova, Vladimir Guzov, Kyle Olszewski, Sergey Tulyakov, and Gerard
  Pons-Moll.
\newblock Control-nerf: Editable feature volumes for scene rendering and
  manipulation.
\newblock {\em arXiv preprint arXiv:2204.10850}, 2022.

\bibitem{legendre2019deeplight}
Chloe LeGendre, Wan-Chun Ma, Graham Fyffe, John Flynn, Laurent Charbonnel, Jay
  Busch, and Paul Debevec.
\newblock Deeplight: Learning illumination for unconstrained mobile mixed
  reality.
\newblock In {\em Proceedings of the IEEE/CVF Conference on Computer Vision and
  Pattern Recognition}, pages 5918--5928, 2019.

\bibitem{lin2021efficient}
Haotong Lin, Sida Peng, Zhen Xu, Hujun Bao, and Xiaowei Zhou.
\newblock Efficient neural radiance fields with learned depth-guided sampling.
\newblock {\em arXiv preprint arXiv:2112.01517}, 2021.

\bibitem{liu2020neural}
Lingjie Liu, Jiatao Gu, Kyaw~Zaw Lin, Tat-Seng Chua, and Christian Theobalt.
\newblock Neural sparse voxel fields.
\newblock {\em NeurIPS}, 2020.

\bibitem{liu2021editing}
Steven Liu, Xiuming Zhang, Zhoutong Zhang, Richard Zhang, Jun-Yan Zhu, and
  Bryan Russell.
\newblock Editing conditional radiance fields.
\newblock In {\em Proceedings of the IEEE/CVF International Conference on
  Computer Vision}, pages 5773--5783, 2021.

\bibitem{marching_cube}
William~E. Lorensen and Harvey~E. Cline.
\newblock Marching cubes: A high resolution 3d surface construction algorithm.
\newblock In {\em Proceedings of the 14th Annual Conference on Computer
  Graphics and Interactive Techniques}, SIGGRAPH '87, page 163–169, New York,
  NY, USA, 1987. Association for Computing Machinery.

\bibitem{marschner1998inverse}
Stephen~Robert Marschner.
\newblock {\em Inverse rendering for computer graphics}.
\newblock Cornell University, 1998.

\bibitem{matusik2003data}
Wojciech Matusik.
\newblock {\em A data-driven reflectance model}.
\newblock PhD thesis, Massachusetts Institute of Technology, 2003.

\bibitem{mildenhall2022nerf}
Ben Mildenhall, Peter Hedman, Ricardo Martin-Brualla, Pratul~P Srinivasan, and
  Jonathan~T Barron.
\newblock Nerf in the dark: High dynamic range view synthesis from noisy raw
  images.
\newblock In {\em Proceedings of the IEEE/CVF Conference on Computer Vision and
  Pattern Recognition}, pages 16190--16199, 2022.

\bibitem{mildenhall2020nerf}
Ben Mildenhall, Pratul~P. Srinivasan, Matthew Tancik, Jonathan~T. Barron, Ravi
  Ramamoorthi, and Ren Ng.
\newblock Nerf: Representing scenes as neural radiance fields for view
  synthesis.
\newblock In {\em ECCV}, 2020.

\bibitem{mueller2022instant}
Thomas M\"uller, Alex Evans, Christoph Schied, and Alexander Keller.
\newblock Instant neural graphics primitives with a multiresolution hash
  encoding.
\newblock {\em ACM Trans. Graph.}, 41(4):102:1--102:15, July 2022.

\bibitem{munkberg2021extracting}
Jacob Munkberg, Jon Hasselgren, Tianchang Shen, Jun Gao, Wenzheng Chen, Alex
  Evans, Thomas M{\"u}ller, and Sanja Fidler.
\newblock Extracting triangular 3d models, materials, and lighting from images.
\newblock {\em arXiv preprint arXiv:2111.12503}, 2021.

\bibitem{oechsle2021unisurf}
Michael Oechsle, Songyou Peng, and Andreas Geiger.
\newblock Unisurf: Unifying neural implicit surfaces and radiance fields for
  multi-view reconstruction.
\newblock In {\em Proceedings of the IEEE/CVF International Conference on
  Computer Vision}, pages 5589--5599, 2021.

\bibitem{ost2022neural}
Julian Ost, Issam Laradji, Alejandro Newell, Yuval Bahat, and Felix Heide.
\newblock Neural point light fields.
\newblock In {\em Proceedings of the IEEE/CVF Conference on Computer Vision and
  Pattern Recognition}, pages 18419--18429, 2022.

\bibitem{park2020seeing}
Jeong~Joon Park, Aleksander Holynski, and Steven~M Seitz.
\newblock Seeing the world in a bag of chips.
\newblock In {\em Proceedings of the IEEE/CVF Conference on Computer Vision and
  Pattern Recognition}, pages 1417--1427, 2020.

\bibitem{qi2017pointnet}
Charles~R Qi, Hao Su, Kaichun Mo, and Leonidas~J Guibas.
\newblock Pointnet: Deep learning on point sets for 3d classification and
  segmentation.
\newblock In {\em Proceedings of the IEEE conference on computer vision and
  pattern recognition}, pages 652--660, 2017.

\bibitem{reiser2021kilonerf}
Christian Reiser, Songyou Peng, Yiyi Liao, and Andreas Geiger.
\newblock Kilonerf: Speeding up neural radiance fields with thousands of tiny
  mlps.
\newblock In {\em Proceedings of the IEEE/CVF International Conference on
  Computer Vision}, pages 14335--14345, 2021.

\bibitem{richter2016instant}
Thomas Richter-Trummer, Denis Kalkofen, Jinwoo Park, and Dieter Schmalstieg.
\newblock Instant mixed reality lighting from casual scanning.
\newblock In {\em 2016 IEEE International Symposium on Mixed and Augmented
  Reality (ISMAR)}, pages 27--36. IEEE, 2016.

\bibitem{schonberger2016pixelwise}
Johannes~L Sch{\"o}nberger, Enliang Zheng, Jan-Michael Frahm, and Marc
  Pollefeys.
\newblock Pixelwise view selection for unstructured multi-view stereo.
\newblock In {\em European conference on computer vision}, pages 501--518.
  Springer, 2016.

\bibitem{sorkine2007rigid}
Olga Sorkine and Marc Alexa.
\newblock As-rigid-as-possible surface modeling.
\newblock In {\em Symposium on Geometry processing}, volume~4, pages 109--116,
  2007.

\bibitem{nerv2021}
Pratul~P. Srinivasan, Boyang Deng, Xiuming Zhang, Matthew Tancik, Ben
  Mildenhall, and Jonathan~T. Barron.
\newblock Nerv: Neural reflectance and visibility fields for relighting and
  view synthesis.
\newblock In {\em CVPR}, 2021.

\bibitem{tancik2020fourier}
Matthew Tancik, Pratul Srinivasan, Ben Mildenhall, Sara Fridovich-Keil, Nithin
  Raghavan, Utkarsh Singhal, Ravi Ramamoorthi, Jonathan Barron, and Ren Ng.
\newblock Fourier features let networks learn high frequency functions in low
  dimensional domains.
\newblock {\em Advances in Neural Information Processing Systems},
  33:7537--7547, 2020.

\bibitem{tang2022compressible}
Jiaxiang Tang, Xiaokang Chen, Jingbo Wang, and Gang Zeng.
\newblock Compressible-composable nerf via rank-residual decomposition.
\newblock {\em arXiv preprint arXiv:2205.14870}, 2022.

\bibitem{tewari2020state}
Ayush Tewari, Ohad Fried, Justus Thies, Vincent Sitzmann, Stephen Lombardi,
  Kalyan Sunkavalli, Ricardo Martin-Brualla, Tomas Simon, Jason Saragih,
  Matthias Nie{\ss}ner, et~al.
\newblock State of the art on neural rendering.
\newblock In {\em Computer Graphics Forum}, volume~39, pages 701--727. Wiley
  Online Library, 2020.

\bibitem{ulyanov2018deep}
Dmitry Ulyanov, Andrea Vedaldi, and Victor Lempitsky.
\newblock Deep image prior.
\newblock In {\em Proceedings of the IEEE conference on computer vision and
  pattern recognition}, pages 9446--9454, 2018.

\bibitem{verbin2021ref}
Dor Verbin, Peter Hedman, Ben Mildenhall, Todd Zickler, Jonathan~T Barron, and
  Pratul~P Srinivasan.
\newblock Ref-nerf: Structured view-dependent appearance for neural radiance
  fields.
\newblock {\em arXiv preprint arXiv:2112.03907}, 2021.

\bibitem{walter2007microfacet}
Bruce Walter, Stephen~R Marschner, Hongsong Li, and Kenneth~E Torrance.
\newblock Microfacet models for refraction through rough surfaces.
\newblock {\em Rendering techniques}, 2007:18th, 2007.

\bibitem{wang2021neus}
Peng Wang, Lingjie Liu, Yuan Liu, Christian Theobalt, Taku Komura, and Wenping
  Wang.
\newblock Neus: Learning neural implicit surfaces by volume rendering for
  multi-view reconstruction.
\newblock {\em NeurIPS}, 2021.

\bibitem{williams1978casting}
Lance Williams.
\newblock Casting curved shadows on curved surfaces.
\newblock In {\em Proceedings of the 5th annual conference on Computer graphics
  and interactive techniques}, pages 270--274, 1978.

\bibitem{xu2022point}
Qiangeng Xu, Zexiang Xu, Julien Philip, Sai Bi, Zhixin Shu, Kalyan Sunkavalli,
  and Ulrich Neumann.
\newblock Point-nerf: Point-based neural radiance fields.
\newblock {\em arXiv preprint arXiv:2201.08845}, 2022.

\bibitem{xu2022deforming}
Tianhan Xu and Tatsuya Harada.
\newblock Deforming radiance fields with cages.
\newblock In {\em ECCV}, 2022.

\bibitem{yang2021learning}
Bangbang Yang, Yinda Zhang, Yinghao Xu, Yijin Li, Han Zhou, Hujun Bao, Guofeng
  Zhang, and Zhaopeng Cui.
\newblock Learning object-compositional neural radiance field for editable
  scene rendering.
\newblock In {\em Proceedings of the IEEE/CVF International Conference on
  Computer Vision}, pages 13779--13788, 2021.

\bibitem{yao2018mvsnet}
Yao Yao, Zixin Luo, Shiwei Li, Tian Fang, and Long Quan.
\newblock Mvsnet: Depth inference for unstructured multi-view stereo.
\newblock {\em European Conference on Computer Vision (ECCV)}, 2018.

\bibitem{yao2020blendedmvs}
Yao Yao, Zixin Luo, Shiwei Li, Jingyang Zhang, Yufan Ren, Lei Zhou, Tian Fang,
  and Long Quan.
\newblock Blendedmvs: A large-scale dataset for generalized multi-view stereo
  networks.
\newblock In {\em Proceedings of the IEEE/CVF Conference on Computer Vision and
  Pattern Recognition}, pages 1790--1799, 2020.

\bibitem{yariv2021volume}
Lior Yariv, Jiatao Gu, Yoni Kasten, and Yaron Lipman.
\newblock Volume rendering of neural implicit surfaces.
\newblock In {\em Thirty-Fifth Conference on Neural Information Processing
  Systems}, 2021.

\bibitem{yifan2020neural}
Wang Yifan, Noam Aigerman, Vladimir~G Kim, Siddhartha Chaudhuri, and Olga
  Sorkine-Hornung.
\newblock Neural cages for detail-preserving 3d deformations.
\newblock In {\em Proceedings of the IEEE/CVF Conference on Computer Vision and
  Pattern Recognition}, pages 75--83, 2020.

\bibitem{Yuan22NeRFEditing}
Yu-Jie Yuan, Yang-Tian Sun, Yu-Kun Lai, Yuewen Ma, Rongfei Jia, and Lin Gao.
\newblock Nerf-editing: Geometry editing of neural radiance fields.
\newblock In {\em Computer Vision and Pattern Recognition (CVPR)}, 2022.

\bibitem{zhang2021physg}
Kai Zhang, Fujun Luan, Qianqian Wang, Kavita Bala, and Noah Snavely.
\newblock Physg: Inverse rendering with spherical gaussians for physics-based
  material editing and relighting.
\newblock In {\em Proceedings of the IEEE/CVF Conference on Computer Vision and
  Pattern Recognition}, pages 5453--5462, 2021.

\bibitem{zhang2020nerf++}
Kai Zhang, Gernot Riegler, Noah Snavely, and Vladlen Koltun.
\newblock Nerf++: Analyzing and improving neural radiance fields.
\newblock {\em arXiv preprint arXiv:2010.07492}, 2020.

\bibitem{zhang2018unreasonable}
Richard Zhang, Phillip Isola, Alexei~A Efros, Eli Shechtman, and Oliver Wang.
\newblock The unreasonable effectiveness of deep features as a perceptual
  metric.
\newblock In {\em Proceedings of the IEEE conference on computer vision and
  pattern recognition}, pages 586--595, 2018.

\bibitem{nerfactor}
Xiuming Zhang, Pratul~P. Srinivasan, Boyang Deng, Paul Debevec, William~T.
  Freeman, and Jonathan~T. Barron.
\newblock Nerfactor: Neural factorization of shape and reflectance under an
  unknown illumination.
\newblock {\em ACM Trans. Graph.}, 40(6), dec 2021.

\end{thebibliography}
